%% file: acl.tex
\pdfoutput=1

\documentclass[11pt]{article}

\usepackage{acl}

\usepackage{times}
\usepackage{latexsym}
\usepackage[T1]{fontenc}

\usepackage[utf8]{inputenc}

\usepackage{microtype}
\usepackage{array}
\usepackage{graphicx}      
\usepackage{booktabs}      
\usepackage{subcaption}    
\usepackage{multirow}      
\usepackage{multicol}      
\usepackage{amssymb}       
\usepackage{pifont}
\newcommand{\cmark}{\ding{51}}
\newcommand{\xmark}{\ding{55}}  
\usepackage{comment}

\title{Large Language Models Are Reasoning Teachers}

\author{\\
   \\
   \\\And
  Namgyu Ho, \,Laura Schmid, \,Se-Young Yun \\
  KAIST \\
  \texttt{\{{itsnamgyu, laura.schmid, yunseyoung\}}@kaist.ac.kr} \\\And
  \\
  \\
  \\}

\begin{document}
\maketitle

\input{main.tex}

\end{document}

%% file: main.tex
\begin{abstract}

Recent works have shown that chain-of-thought (CoT) prompting can elicit language models to solve complex reasoning tasks, step-by-step.
However, prompt-based CoT methods are dependent on very large models such as GPT-3 175B which are prohibitive to deploy at scale.
In this paper, we use these large models as \textit{reasoning teachers} to enable complex reasoning in smaller models and reduce model size requirements by several orders of magnitude.
We propose \textit{Fine-tune-CoT}, a method that generates reasoning samples from very large teacher models to fine-tune smaller models.
We evaluate our method on a wide range of public models and complex tasks.
We find that Fine-tune-CoT enables substantial reasoning capability in small models, far outperforming prompt-based baselines and even the teacher model in many tasks.
Additionally, we extend our method by leveraging the teacher model's ability to generate multiple distinct rationales for each original sample.
Enriching the fine-tuning data with such \textit{diverse reasoning} results in a substantial performance boost across datasets, even for very small models.
We conduct ablations and sample studies to understand the emergence of reasoning capabilities of student models.\footnote{
Our code implementation and data are available at \href{https://github.com/itsnamgyu/reasoning-teacher}{https://github.com/itsnamgyu/reasoning-teacher}.
}
%
%
%

\end{abstract}

\input{section1_introduction}

\input{section2_related}
\input{section3_method}
\input{section4_experiments}
\input{section5_discussion}

\input{section6_conclusion}
\input{section7_limitations}

\section*{Acknowledgements}

This work was supported by Institute of Information \& communications Technology Planning \& Evaluation (IITP) grant funded by Korea government (MSIT) [No. 2021-0-00907, Development of Adaptive and Lightweight Edge-Collaborative Analysis Technology for Enabling Proactively Immediate Response and Rapid Learning, 90\%], [No. 2019-0-00075, Artificial Intelligence Graduate School Program (KAIST), 5\%], and the Stochastic Analysis and Application Research Center (SAARC) under the National Research Foundation of Korea grant (NRF-2019R1A5A1028324, 5\%).

\bibliography{anthology,custom}
\bibliographystyle{acl_natbib}

\clearpage

\appendix

\input{appendix}

%% file: section1_introduction.tex
\section{Introduction}
\label{introduction}

\begin{figure}[t]
    \centering
    \resizebox{1\linewidth}{!}{
        \includegraphics{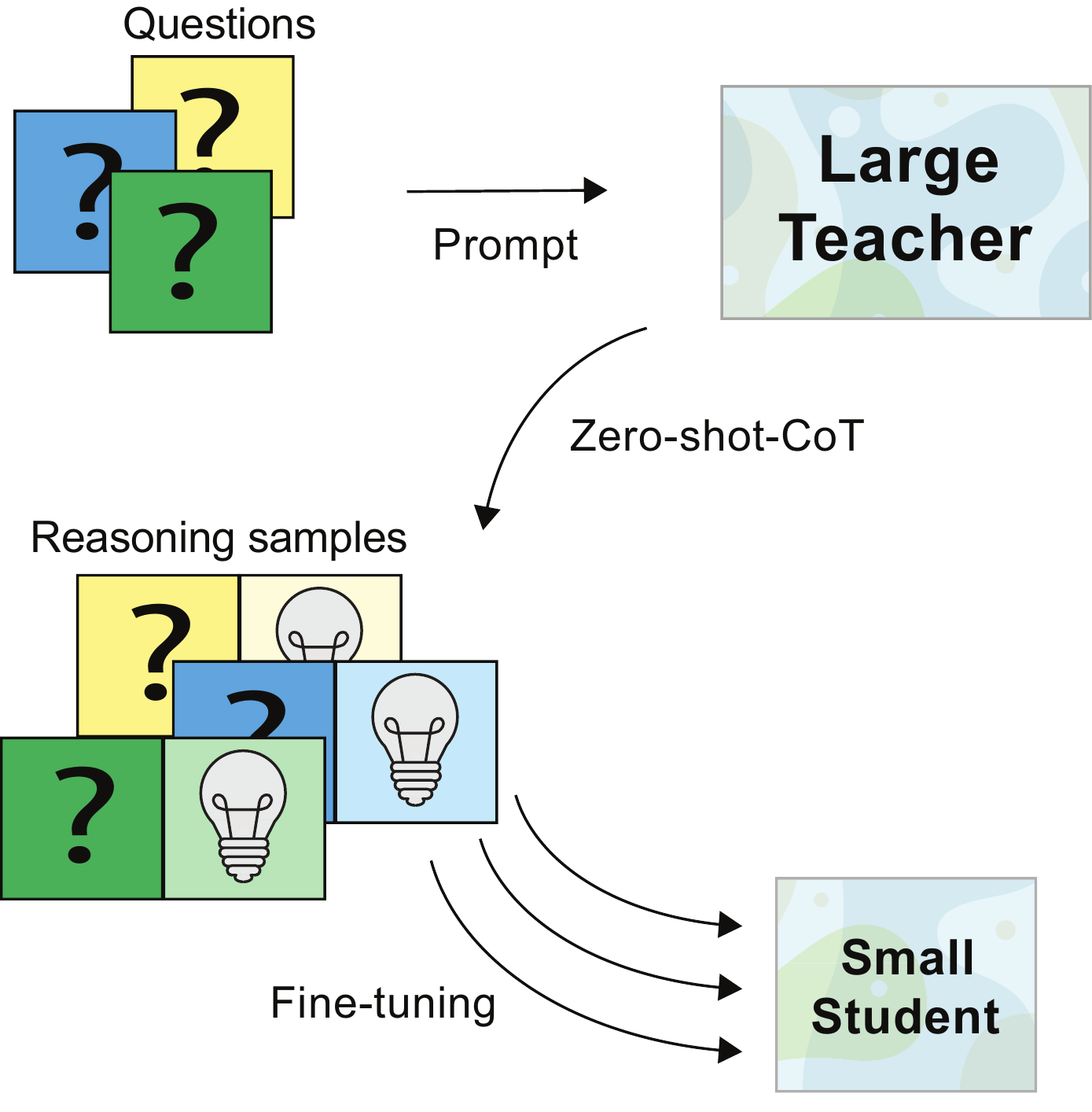}
    }
    \caption{
    \textbf{Fine-tune-CoT uses teacher-generated reasoning to teach students}.
    We prompt a very large teacher model, such as GPT-3 175B, to solve complex questions via zero-shot chain-of-thought reasoning. We then use the reasoning samples to fine-tune a much smaller student model. See Figure \ref{fig:overview} for details.
    }
    \label{fig:scalability}
\end{figure}

\begin{figure*}
    \centering
    \resizebox{\linewidth}{!}{
        \includegraphics{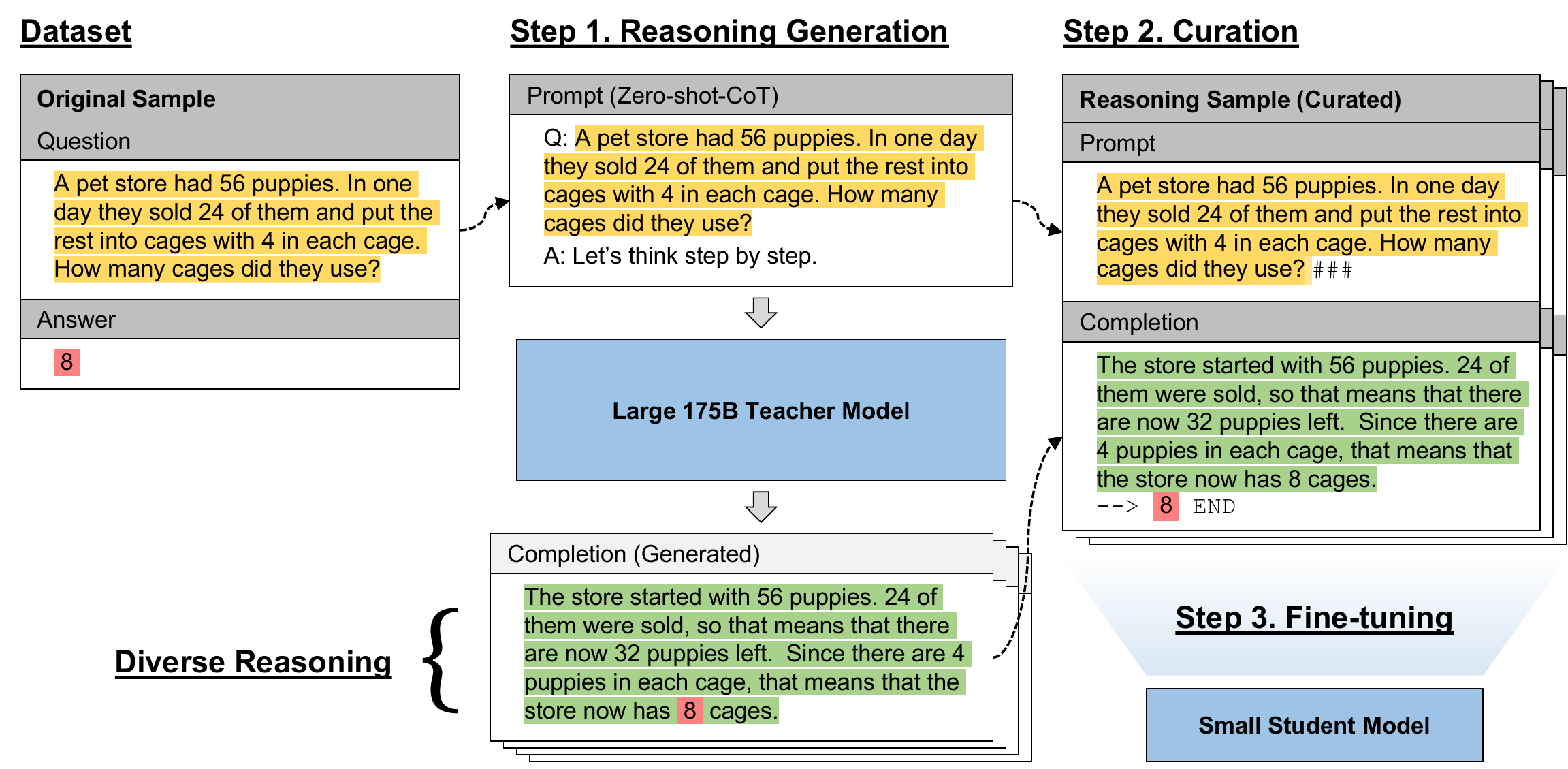}
    }
    \caption{
    Detailed overview of our proposed Fine-tune-CoT method.
    \textbf{Step 1}: a very large teacher model is prompted to solve complex questions (yellow) by generating multi-step reasoning explanations (green).
    \textbf{Step 2}: completions are filtered based on the correctness of the final prediction (red).
    The question, rationale, and answer are used to compose a \textit{reasoning sample} comprised of the prompt and a multi-step solution.
    \textbf{Step 3}: the curated reasoning samples are used to fine-tune a small, lightweight student to exhibit reasoning capabilities.
    The application of an \textit{LM-based} teacher enables \textbf{diverse reasoning}---generating multiple distinct rationales for each original sample to enrich the fine-tuning data.
    This boosts the performance of student models without any additional human annotation.
    }
    \label{fig:overview}
\end{figure*}

Language models (LMs) have demonstrated remarkable performance in a wide range of downstream tasks.
Recently, large language models (LLMs) have demonstrated in-context generalization capabilities: performing downstream tasks simply by conditioning on few in-context exemplars or plain natural language task descriptions \citep{gpt3__brown2020language,sun2021ernie}.
Despite these advancements, even the largest LLMs have been found to struggle with complex tasks which require multiple reasoning steps \citep{gopher__rae2021scaling}.

To solve complex tasks, recent works show that it is possible to elicit reasoning abilities by prompting LLMs to perform \textit{chain-of-thought} (CoT) reasoning, i.e., generate a series of intermediate reasoning steps.
This can be achieved by providing CoT demonstrations as exemplars in prompting \citep{wei2022chain}.
More recently, \citet{kojima2022large} found that LLMs can be prompted to perform CoT reasoning simply by providing a natural language instruction to \textit{think step-by-step}.
%
%

A major drawback of prompt-based CoT reasoning methods, however, is their reliance on extremely large models that span \textit{hundreds of billions} of parameters \citep{wei2022chain, kojima2022large}.
These models are prohibitive to deploy at scale due to overwhelming computational requirements and inference costs \citep{wei2022chain}.
Therefore, we strive to enable complex reasoning in small models which are more feasible for large-scale deployment.

In this light, we propose an approach named \emph{Fine-tune-CoT}, which utilizes the reasoning capabilities of very large LMs to teach small models how to solve complex tasks.
We apply existing zero-shot CoT prompting \citep{kojima2022large} to generate rationales from very large \textit{teacher} models, and use them to fine-tune smaller \textit{student} models\footnote{This can be interpreted as a variant of knowledge distillation~\citep{hinton2015distilling}.}. We illustrate this in Figure \ref{fig:overview}.
We note that standard fine-tuning \textit{without rationales} has been shown to be inadequate for solving reasoning tasks with small models \citep{talmor2018commonsenseqa}.
While there have been attempts to fine-tune small models with \textit{hand-annotated} reasoning steps \citep{nye2021scratchpad, gsm8k__cobbe2021training}, they often require task-specific training setups and high-quality rationales which are costly to annotate \citep{wei2022chain}.  
In contrast, our approach can be readily applied to novel downstream tasks 
without hand-crafted reasoning or task engineering.

We also propose a novel extension to our method, termed \textit{diverse reasoning}, to maximize the teaching effects of Fine-tune-CoT. Inspired by the intuition that complex tasks can have multiple solutions with distinct reasoning paths \citep{evans2010intuition}, we generate multiple reasoning solutions from teacher models using stochastic sampling to augment the training data for student models\footnote{Diverse reasoning is orthogonal to existing data augmentation techniques \citep{yoo2021gpt3mix, meng2022generating} which aim to augment new \textit{question-answer} pairs rather than diverse reasoning \textit{solutions} for complex questions.}.
We find that this is a simple yet highly effective approach to maximizing student performance, which has not been explicitly recognized in concurrent works on fine-tuning with CoT reasoning \citep{huang2022large, li2022diverse, magister2022teaching, fu2023specializing}.

We evaluate our method on 12 tasks using a wide range of publicly available models.
We find that Fine-tune-CoT can elicit notable reasoning performance in small models while preserving much of the versatility of prompt-based CoT reasoning, which previously required $>$100B parameter models \citep{wei2022chain}.
Diverse reasoning enables remarkable gains in performance at the minor cost of additional teacher inference at development time, by exploiting our unique learning setup.
This enables models as small as 0.3B to outperform larger students, and even the 175B teacher model in some tasks.
Our ablations show that performance is consistently scalable across all axes considered: diverse reasoning, dataset size, teacher performance, and student model size.
This shows the potential of our method to enable reliable performance in small models that are feasible for use in real-world applications.
Lastly, we conduct thorough sample studies and analyses which shed light on crucial details previous overlooked in fine-tuning for CoT and provide intuition on the emergence of reasoning abilities in small models.

%% file: section2_related.tex
\section{Related Work}
\label{section:related}


\paragraph{Downstream transfer in language models}
Much previous work established a ``pre-train and fine-tune'' 
paradigm for enhancing LLM performance on downstream tasks~\cite{gpt1__radford2018improving,dong2019pretrain,vaswani2017attention,devlin2018bert}.
However, 
fine-tuning is not always easily applicable~\citep{hendrycks2019ood}.
More recent literature exhibits a paradigm shift towards ``prompting'' the model to predict the desired output~\citep{liu2021prompt,t5__raffel2020exploring}. 
Large LMs can exhibit strong performance in this setting~\citep{gpt3__brown2020language}. For smaller models to be able to perform similarly, additional engineering is usually required~\cite{gao-etal-2021-making,schick-schutze-2021-just,schick-etal-2020-automatically}. For more complex tasks, the idea of using samples with explicit reasoning steps for fine-tuning a model~\citep{nye2021scratchpad, gsm8k__cobbe2021training} preceded the approach of
chain-of-thought (CoT) prompting~\citep{wei2022chain}, which enables very large LMs to perform well.

\paragraph{Chain-of-thought reasoning}
In few-shot CoT prompting, the model learns to generate intermediate reasoning steps that lead to a problem solution, after being fed examples of step-by-step reasoning. 
This enables very good performance on a wide range of tasks. 
~\citep{wang2022self}. Additionally, LLMs can perform well in an unsupervised task-agnostic setting, using Zero-shot-CoT~\citep{kojima2022large}. 
This requires no fine-tuning or task specific conditioning, 
and substantially outperforms standard zero-shot learning and sometimes even few-shot learning on a wide number of tasks. 

Yet, prior work has shown that CoT 
requires extremely large models for optimal performance~\citep{hoffmann2022training,chowdhery2022palm}.
In our work, we contrast this by showing how to utilize CoT reasoning methods for smaller models by fine-tuning them on rationales generated by a very large model. 
Using various LLM-generated explanations for fine-tuning smaller models has been successfully used in prior work~\citep{li2022explanations}, with a focus on specific single tasks. Also, a similar approach to ours is mentioned in~\citep{huang2022large}; however we note that this concurrent work focuses on using Few-shot-CoT to self-generate fine-tuning examples by and for very large proprietary models.
There is a brief glimpse into fine-tuning on smaller distilled models, but the results are limited to one dataset and very large teacher models that are inaccessible to the general community.
In contrast, we provide a rich set of results and qualitative/quantitative analysis on a wide range of datasets, using open-source models that are small and accessible to everyone.





\paragraph{Knowledge distillation}

Typically, knowledge distillation (KD) refers to training small models derived from large models in order to reduce model size and latency, while still preserving accuracy and capacity to generalize~\citep{hinton2015distilling, sanh2019distilbert}. Essentially, KD is a form of model compression, making efficient deployment to capacity-limited devices possible~\citep{bucilua2006compression}. We note that our work could also be considered a distant variant of KD~\citep{Gou2021kdsurvey}, similar to works on improving prompt-based methods such as \citet{yoo2021gpt3mix, schick-schutze-2021-just,schick-schutze-2021-exploiting,zelikman2022star}, or works on data-free distillation~\citep{micaelli2019zero,nayak2019zero,shen2021progressive}, where the transfer data is synthetically generated from a large teacher model. 
Similarly, sequence-level distillation, i.e. training a student model on sequence distributions of a larger teacher, can make neural machine translation more efficient~\citep{kim2016sequence}. 
Despite being similar in spirit, our method still distinguishes itself from such previous work. The role of the teacher model in our method is to teach the notion of intermediate reasoning. It is not the specific output that is the main supervising signal for reasoning, but rather the generation's structure. Hence, we do not use a standard KD loss function that reflects trying to match the teacher output.  
Adding to this, we note that our diverse reasoning is also unusual in the context of KD, where it is e.g. sufficient in practice to only generate one teacher sequence for sequence level distillation.

%% file: section3_method.tex
\section{Chain-of-Thought Fine-Tuning}
\label{method}

We propose Fine-tune-CoT, a task-agnostic approach to enable chain-of-thought reasoning in small LMs.
The core idea is to generate reasoning samples from very large teacher models using CoT prompting and subsequently fine-tune small student models using the generated samples.
This approach preserves the versatility of prompt-based CoT methods while overcoming their reliance on prohibitively large models.
To maximize versatility and minimize teacher inference costs, we use the task-agnostic Zero-shot-CoT prompting method \citep{kojima2022large} on teacher models, as it does not require \textit{any} reasoning examples or long inference context.
We discuss our choice of teacher CoT prompting method in Section~\ref{cot_methods}.
In the following, we characterize Fine-tune-CoT in three distinct steps. We also provide a visual overview in Figure~\ref{fig:overview}.

\paragraph{Step 1. Reasoning generation} First, we utilize a large teacher model to generate CoT reasoning explanations for a given task. Consider a standard sample $S_i$ consisting of a question $q_i$ and its true answer $a_i$. Using Zero-shot-CoT\,\footnotemark.
we prompt the teacher model to generate a reasoning explanation, or rationale, $\hat{r}_i$ to solve question $q_i$ and make a final answer prediction $\hat{a}_i$.
The resulting text sequence, including the prompt and generations, takes the following form: ``Q: <$q_i$>. A: Let's think step by step. \textcolor{blue}{<$\hat{r}_i$>} Therefore, the answer is \textcolor{red}{<$\hat{a}_i$>}''.

\paragraph{Step 2. Curation} Next, we filter the generated samples and reformat them into prompt-completion pairs.
For filtering, we simply compare the final prediction of the teacher model $\hat{a}_i$ with the ground-truth answer $a_i$, following previous works~\citep{zelikman2022star, huang2022large}.
Note that this filtering incurs some loss of training samples.
For all instances $i$ where $\hat{a}_i = a_i$, we repackage $(S_i, \hat{r}_i, \hat{a}_i)$ into a \textit{reasoning sample} $S^\prime_i = (p_i, c_i)$, a prompt-completion pair.
To maximize inference-time efficiency, we use special-character based delimiters to minimize token usage.
Specifically, $p_i$ and $c_i$ each take the form of ``<$q_i$> \texttt{\#\#\#}'' and ``<$\hat{r}_i$> \texttt{{-}{-}{>}} <$a_i$> \texttt{END}''.
We note that answer-based filtering does not ensure the correctness of the rationales, especially for multi-choice questions.
We provide an analysis in Appendix~\ref{rationale_filtering} regarding this important detail which has not been addressed in concurrent work.



\paragraph{Step 3. Fine-tune}
Finally, we fine-tune a small pre-trained student model on the assembled reasoning samples.
We use the same training objective of that used during pre-training, i.e., autoregressive language modeling objective, or next-token prediction \citep{gpt1__radford2018improving}.

\footnotetext{Note that Zero-shot-CoT is itself a two-step prompting method. The reasoning (blue) is generated in the first step and answer prediction (red) is generated in the second step.}

\paragraph{Diverse reasoning}
To maximize the teaching effects of Fine-tune-CoT, we can generate multiple reasoning explanations for each training sample.
This approach is motivated by the intuition that multiple reasoning paths can be used to solve complex tasks, i.e., \textit{type-2} tasks \citep{evans2010intuition}.
We posit that this unique feature of complex tasks, in tandem with the stochastic generation abilities of the teacher model, can enable diverse reasoning to significantly boost reasoning supervision simply through additional teacher inference.
In detail, for a given sample $S_i$, instead of applying Zero-shot-CoT using greedy decoding to obtain a single explanation-answer pair $(\hat{e}_i, \hat{a}_i)$, we use a stochastic sampling strategy, i.e., temperature sampling with large $T$, to obtain $D$ distinct generations $\{(\hat{r}_{ij}, \hat{a}_{ij})\}^{D}_{j}$.
Subsequent reasoning sample curation and fine-tuning then proceed as before.
We refer to $D$ as the \textit{degree of reasoning diversity}.
A similar approach is used in \citet{wang2022self, huang2022large}, where multiple CoT outputs are generated and marginalized to find the optimal answer.
However, the effects of such diverse reasoning on teaching student models has not been acknowledged or thoroughly investigated in concurrent work \citep{huang2022large, li2022explanations, magister2022teaching, fu2023specializing}.
We note that diverse reasoning imposes an important tradeoff between the development cost and inference cost/quality of student models which we discuss in Section~\ref{tradeoffs}.

%% file: section4_experiments.tex
\begin{table*}[h]
\begin{minipage}[b]{1.0\linewidth}
\vspace*{0.25cm}
\centering
\resizebox{\linewidth}{!}{
\begin{tabular}{llrrrrrrrrrrrrr}
\toprule
\multirow[c]{2}{*}{Method} & \multirow[c]{2}{*}{Params} & Single & Add & Multi & \multirow[c]{2}{*}{GSM8K} & \multirow[c]{2}{*}{Aqua} & \multirow[c]{2}{*}{SVAMP} & Date & Shuffled & Last & Coin & Common & Strategy \\
& & Eq & Sub & Arith &  &  &  & Understanding & Objects & Letter & Flip & SenseQA & QA  \\
\midrule
 Random &  & 0.00 & 0.00 & 0.00 & 0.00 & 20.00 & 0.00 & 17.12 & 33.33 & 0.00 & 50.00 & 20.00 & 50.00 \\
\midrule
\multicolumn{14}{c}{\textbf{Teacher: InstructGPT (\texttt{text-davinci-002})}} \\
\midrule
Zero-shot-CoT & 175B & 82.24 & 78.99 & 78.89 & 40.26 & 34.25 & 64.67 & 73.87 & 50.22 & 56.00 & 92.67 & 61.75 & 53.57 \\
\midrule
\multicolumn{14}{c}{\textbf{Student: GPT-3 (\texttt{ada}, \texttt{babbage}, \texttt{curie})}} \\
\midrule
Zero-shot & 6.7B & 0.66 & 0.84 & 3.33 & 1.74 & 16.54 & 2.67 & 9.91 & 32.89 & 0.00 & 56.67 & 20.23 & 52.98 \\
\midrule
Zero-shot-CoT & 6.7B & 1.32 & 2.52 & 5.00 & 2.35 & 21.26 & 1.33 & 15.32 & 31.11 & 0.00 & 46.67 & 19.98 & 51.09 \\
Few-shot-CoT & 6.7B & 22.37 & \textbf{31.93} & 10.00 & 2.50 & 15.75 & 11.33 & 12.84 & - & 0.67 & 40.00 & 24.73 & 54.68 \\
\midrule
Fine-tune & 6.7B & \textbf{24.34} & 25.21 & 15.00 & 6.14 & 15.35 & 20.67 & 14.41 & 33.78 & 32.67 & 72.00 & \textbf{76.17} & \textbf{65.21} \\
\midrule
Fine-tune-CoT & 0.3B & 7.24 & 6.72 & 6.11 & 3.11 & 23.62 & 5.00 & 17.12 & 49.33 & 50.67 & 99.33 & 32.68 & 52.55 \\
 & 1.3B & 11.18 & 11.76 & 13.33 & 4.70 & 19.69 & 8.00 & 38.74 & 52.44 & 50.67 & 100.00 & 43.08 & 52.69 \\
  & 6.7B & 20.39 & 21.01 & 33.33 & \textbf{6.75} & \textbf{24.02} & 12.67 & 60.36 & 64.44 & 52.67 & 98.67 & 56.76 & 55.02 \\
\midrule
Fine-tune-CoT  & 0.3B & 9.21 & 10.08 & 23.89 & - & - & 14.33 & 58.56 & 61.78 & 59.33 & 99.33 & - & 57.21 \\
\small{w/ diverse reasoning} & 1.3B & 18.42 & 19.33 & 27.78 & - & - & 16.33 & 70.27 & 72.00 & 60.67 & 100.00 & - & 57.06 \\
 & 6.7B & \textbf{24.34} & 31.09 & \textbf{53.33} & - & - & \textbf{30.33} & \textbf{83.78} & \textbf{73.33} & \textbf{62.00} & \textbf{100.00} & - & 58.22 \\
\bottomrule
\end{tabular}
}
\end{minipage}
\caption{
    \textbf{Fine-tune-CoT Performance.} Accuracy (\%) of OpenAI models on 12 tasks under Fine-tune-CoT (with diverse reasoning) and baseline methods.
   `Random' refers to random-guess performance derived based on the number of choices in multi-choice tasks.
   For diverse reasoning, we report results for maximum degree $D$ considered: $D=64$ for MultiArith and SVAMP; $D=8$ for other datasets.
   We omit diverse reasoning for large datasets due to resource constraints and Few-shot-CoT for Tracking Shuffled Objects due to absence of prompts.
}
\label{tab:performance}
\end{table*}

\section{Experiments}
\label{experiments}

\paragraph{Tasks and datasets}
We evaluate our method on 12 datasets pertaining to four categories of complex reasoning, following \citet{kojima2022large}. These include arithmetic (SingleEq, AddSub, MultiArith, GSM8K, SVAMP), other (Date Understanding, Tracking Shuffled Objects), symbolic (Last Letter Concatenation, Coin Flip), and common sense (CommonSenseQA, StrategyQA) reasoning.
We provide details and references in Appendix~\ref{datasets}.

\paragraph{Models}
For teacher models, we use four variants of GPT-3 175B \citep{gpt3__brown2020language}, provided by the OpenAI API.
Unless otherwise stated, we use \texttt{text-davinci-002} based on InstructGPT 175B \citep{ouyang2022instruct} as the teacher for Fine-tune-CoT.
For student models, we consider four popular model families. For our main experiments, we use GPT-3 \{\texttt{ada}, \texttt{babbage}, \texttt{curie}\} as they are readily available for fine-tuning via the OpenAI API. Due to the blackbox nature of the API, we also consider various open-source models under controlled settings.
We use GPT-2 \{Small, Medium, Large\} \citep{gpt2__radford2019language} and T5-\{Small, Base, Large\} \citep{t5__raffel2020exploring} as representative model families for decoder-only and encoder-decoder architectures, respectively.
We also use the instruction-tuned version of T5, Flan-T5-\{Small, Base, Large\} \citep{flan_t5__chung2022scaling}, to investigate the effects of instruction tuning on student models, prior to applying Fine-tune-CoT.
These student models are 25--2500x smaller than the teacher model, thus considerably more feasible for real-world deployment.
We provide details on models and API usage in Appendix~\ref{models}.

\begin{table}[t]
\resizebox{\linewidth}{!}{
\begin{tabular}{lccccl}
\toprule
\multirow[c]{2}{*}{Method} & Model & CoT & Sample & Teacher & \multirow[c]{2}{*}{Reference} \\
& Updates & Output & Utilization & Usage & \\
\midrule
Zero-shot & \xmark & \xmark & \xmark & \xmark & \cite{gpt2__radford2019language} \\
Zero-shot-CoT & \xmark & \cmark & \xmark & \xmark & \cite{kojima2022large} \\
Few-shot-CoT & \xmark & \cmark & $\triangle$ & \xmark & \cite{wei2022chain} \\
Fine-tune & \cmark & \xmark & \cmark & \xmark & \cite{gpt1__radford2018improving} \\
Fine-tune-CoT & \cmark & \cmark & \cmark & \cmark & Ours \\
\bottomrule
\end{tabular}
}
\caption{\textbf{Taxonomy of methods}. CoT methods are more interpretable due to reasoning output. While Few-shot-CoT can utilize few in-context samples, fine-tuning can utilize any number of training samples via model updates. Fine-tune-CoT benefits from the reasoning capabilities of teacher models.
}
\label{tab:methods}
\end{table}

\paragraph{Baseline methods}
We provide a comparison of Fine-tune-CoT (ours) with four baseline methods: standard zero-shot prompting, vanilla fine-tuning, Zero-shot-CoT \citep{kojima2022large}, and Few-shot-CoT \citep{wei2022chain}.
Given a training sample $\{(q_i, a_i)\}_i$, we use a simple format ``Q: <$q_i$>'' for zero-shot prompting.
For vanilla fine-tuning, we format the prompt and completion as ``<$q_i$> \texttt{\#\#\#}'' and ``<$a_i$> \texttt{END}'', respectively.
We clarify the taxonomy of methods in Table~\ref{tab:methods}.
For text generation, we use greedy decoding following \citet{wei2022chain, kojima2022large} throughout our experiments, except for diverse reasoning. For diverse reasoning on the teacher, we use temperature sampling with $T=0.7$, following \citet{wang2022self}.
We provide experimental details in Appendix~\ref{experimental_details}.

\subsection{Results}
\label{results}

\begin{figure}[h]
    \centering
    \resizebox{\linewidth}{!}{
        \includegraphics{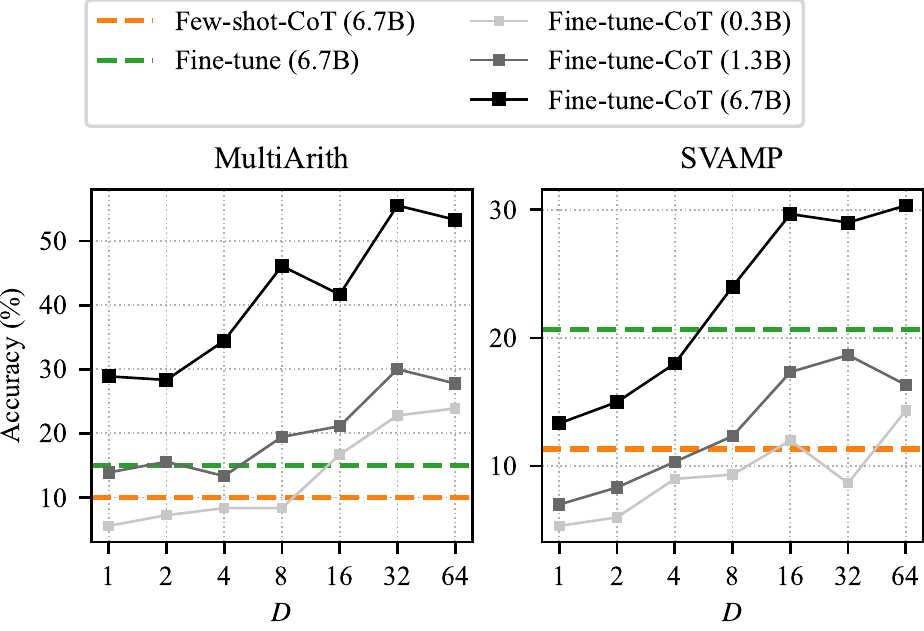}
    }
    \caption{
    \textbf{Diverse reasoning performance.} Accuracy (\%) of GPT-3 student models under Fine-tune-CoT with varying degrees of diverse reasoning $D$. Baseline performance of the \textit{largest model} under vanilla fine-tuning and Few-shot-CoT are shown for comparison. Diverse reasoning is not applicable to the baselines.
    }
    \label{fig:diverse_reasoning}
\end{figure}

\begin{figure}[ht]
    \centering
    \resizebox{\linewidth}{!}{
        \includegraphics{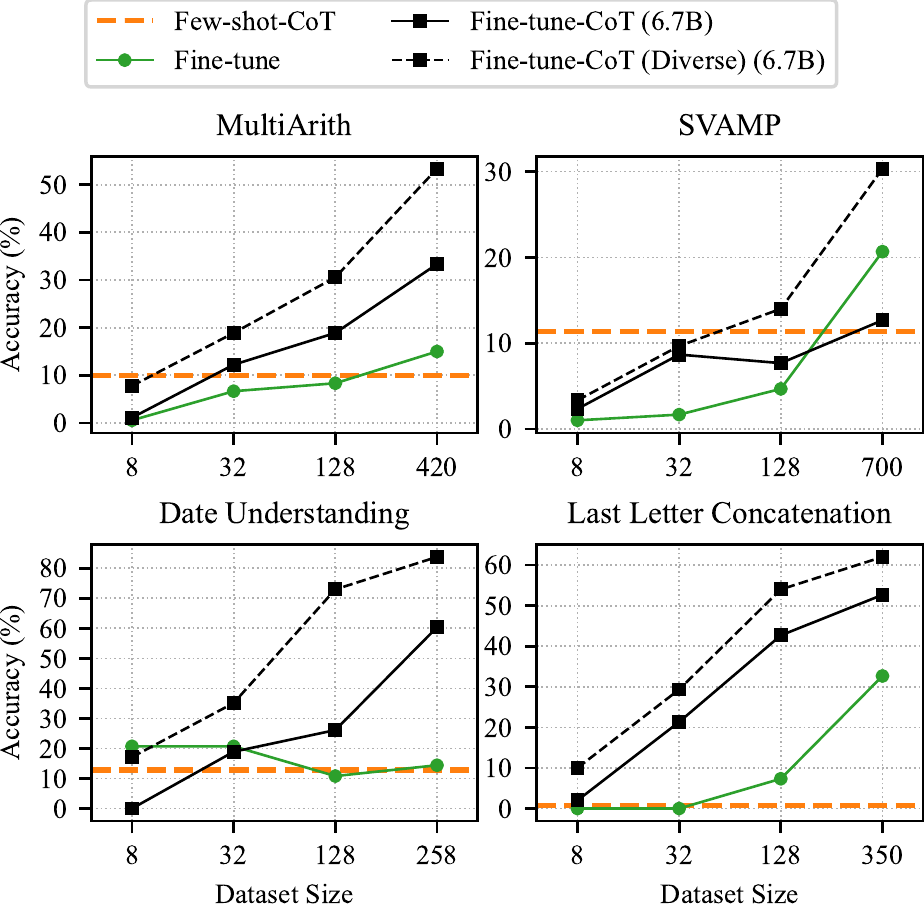}
    }
    \caption{
    \textbf{Effects of dataset size.} 
    Accuracy (\%) of the GPT-3 6.7B student model by dataset size under vanilla fine-tuning vs Fine-tune-CoT (with diverse reasoning). Baseline performance under Few-shot-CoT is shown for comparison. Diverse reasoning is not applicable to standard fine-tuning. We show diverse reasoning performance with $D=64$ for MultiArith and SVAMP; $D=8$ for others.
    }
    \label{fig:dataset_size}
\end{figure}

\begin{figure}[ht]
    \centering
    \resizebox{\linewidth}{!}{
        \includegraphics{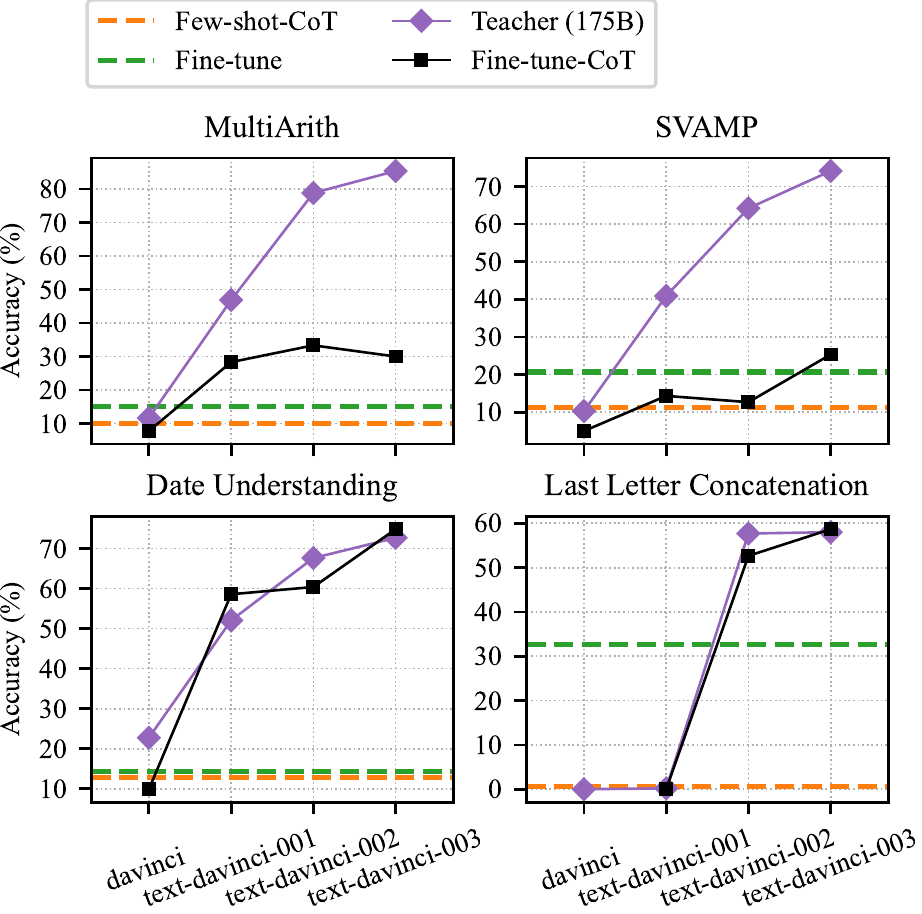}
    }
    \caption{
    \textbf{Effects of teacher performance on students.}
    Accuracy (\%) of teacher models (Zero-shot-CoT) and their corresponding GPT-3 6.7B student models (Fine-tune-CoT).
    Baseline performance under vanilla fine-tuning and Few-shot-CoT are shown for comparison. 
    Teacher models are not applicable to Few-shot-CoT which uses few human-annotated examples.
    }
    \label{fig:teacher_performance}
\end{figure}

\begin{figure}[h!]
    \centering
    \includegraphics[width=\linewidth]{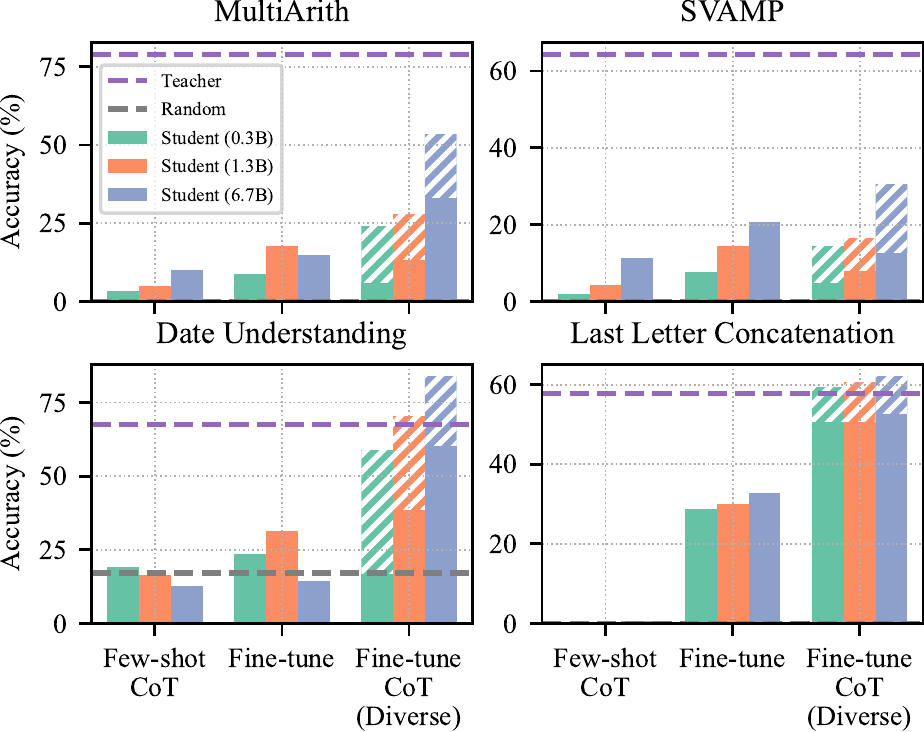}
    \caption{
    \textbf{Effects of student model scale.}
    Accuracy (\%) of GPT-3 student models of various sizes under Few-shot-CoT, vanilla fine-tuning, and Fine-tune-CoT (with diverse reasoning). The hatched portion indicates the performance boost of Fine-tune-CoT when using diverse reasoning with $D=64$ on MultiArith and SVAMP; $D=8$ on others.
    }
    \label{fig:student_size}
\end{figure}

In this section, we present the reasoning performance of models using Fine-tune-CoT and diverse reasoning.
We compare with various baselines and demonstrate the scalability of our method across four axes: degree of diverse reasoning (Figure~\ref{fig:diverse_reasoning}), dataset size (Figure~\ref{fig:dataset_size}), performance of the teacher (Figure~\ref{fig:teacher_performance}), and size of the student model (Figure~\ref{fig:student_size}). \linebreak
We present our findings on GPT-3 models in the main text and defer results on open-source models to Appendix~\ref{open_source_experiments}, with a brief summary at the end of this section.



\paragraph{Fine-tune-CoT elicits complex reasoning in small models}
Table~\ref{tab:performance} summarizes the accuracy of student models using the proposed Fine-tune-CoT, compared to prompt-based CoT baselines as well as standard fine-tuning.
While Zero-shot-CoT exhibits remarkable performance on the very large 175B model \citep{kojima2022large}, it fails to enable complex reasoning in all three smaller models, showing near-negligible performance across \textit{all} tasks.
We also find that small models are unable to approach these tasks under standard zero-shot prompting.
On the other hand, Fine-tune-CoT elicits notable reasoning performance, demonstrating significant gains over Zero-shot-CoT when using smaller models and outperforming both fine-tuning and Few-shot-CoT in more than half of the tasks.
For complex arithmetic, Fine-tune-CoT achieves a notable 33\% accuracy on MultiArith while Zero-shot-CoT only reaches 5\%. Few-shot-CoT and fine-tuning only achieve 10\% and 15\%, respectively. 
For two commonsense reasoning tasks, our method outperforms the near-random performance of Zero-shot-CoT by 37\% and 5\%, respectively. Furthermore, it surpasses Few-shot-CoT on CommonSenseQA by 32\% and performs similarly on StrategyQA.
We observe that Fine-tune-CoT performance is most notable for tasks that are not overly complex, which include other reasoning tasks (Date Understanding, Shuffled Objects) and symbolic reasoning (Last Letter, Coin Flip), significantly outperforming other baselines.
See Appendix Table \ref{tab:performance_all} for performance of all students.

\paragraph{Small models can outperform very large teachers in reasoning}
Table~\ref{tab:performance} also shows that Fine-tune-CoT is highly effective on small models compared to the large 175B teacher model.
For the tasks Shuffled Objects and Coin Flip, Fine-tune-CoT is shown to outperform the teacher model using either 1.3B or 6.7B parameters, i.e., reducing the number of required parameters by approx. 25--100x.
We also find that Fine-tune-CoT with the very small 0.3B model consistently outperforms the 6.7B model under Zero-shot-CoT, demonstrating that our method is able to unlock a wider range of capabilities compared to the baseline, even when model size is vastly reduced.

\paragraph{Diverse reasoning substantially improves Fine-tune-CoT performance.}
To examine the learning effects of diverse reasoning and compare it with two baselines given by fine-tuning and Few-shot-CoT, we apply Fine-tune-CoT using 1--64 reasoning explanations per sample across three model scales on MultiArith and SVAMP\footnote{
For diverse reasoning, we generate teacher rationales stochastically with $T=0.7$ instead of greedy decoding, which accounts for small differences in absolute performance numbers between Table~\ref{tab:performance} and diverse reasoning with $D=1$. 
}.
Figure~\ref{fig:diverse_reasoning} shows that diverse reasoning can significantly improve the performance of student models using Fine-tune-CoT. For the 6.7B student model, we find a boost of around 26\% on MultiArith, and around 17\%  on SVAMP. We also note that using diverse reasoning always leads to outperforming the baseline within the respective model size, and can even boost performance of our method beyond that of a larger model that does not use diverse reasoning.  This even includes the teacher in two cases (Date Understanding, Last Letter).
Moreover, we find that diverse reasoning can boost the performance of Fine-tune-CoT to surpass that of both Few-shot-CoT and vanilla fine-tuning across \emph{all} model sizes.
We posit that due to our focus on \textit{complex tasks}, the diversity of reasoning paths and linguistic templates can substantially aid in teaching student models to reason.

\paragraph{Fine-tune-CoT consistently benefits from more data.}
We perform an ablation on dataset size to study the performance scalability of our method with dataset size.
We see that the performance of the 6.7B model clearly scales with the size of the dataset, independent of the task. In comparison, vanilla fine-tuning does not always exhibit this behavior. In fact, for Date Understanding, we find that an increase in dataset size harms the performance of fine-tuning.
Furthermore, Fine-tune-CoT sees additional benefits from diverse reasoning, which is not applicable in standard fine-tuning.

\paragraph{Better reasoners are better teachers}
Next, we can ask the question of whether the performance of the teacher is correlated with that of their student when using Fine-tune-CoT. To test this, we use different versions of GPT-3 as teacher models, keeping the size of the student model constant at 6.7B parameters (Figure~\ref{fig:teacher_performance}). We find that student performance indeed scales with teacher performance, particularly in the less complex tasks Date Understanding and Last Letter. There, the performance of the student matches the performance of the teacher very closely. This also fits with our observations in Appendix~\ref{sample_study}, which show that the successes and failures of teachers are correlated with those of the students. We note that this scaling effect is in contrast not a given in knowledge distillation, where more accurate teachers do not always result in better students~\cite{menon2021statkd}.

\paragraph{Fine-tune-CoT performance scales with model size for small LMs}
Finally, we explore the effect of scaling up student model size on our method, and compare it with the effects of increasingly larger student models in Few-shot-CoT as well as vanilla fine-tuning. We can observe that the performance of Fine-tune-CoT is consistently scalable with student size (Figure~\ref{fig:student_size}). In contrast, the two baselines do not always exhibit the same behavior: in Date Understanding, neither Few-shot-CoT nor vanilla fine-tuning results in scalable performance.

\paragraph{Results on open-source student models}
Overall, our findings on T5, Flan-T5, and GPT-2 show similar trends to those observed on GPT-3. Small models exhibit near-random performance under standard zero-shot or CoT prompting in nearly all cases.
Notable, we find that encoder-decoder models, i.e., T5 and Flan-T5, show noteworthy performance under standard fine-tuning, suggesting that causal masking may be a bottleneck to reasoning in decoder-based language models in the absence of CoT output.
Fine-tune-CoT consistently outperforms prompt-based baselines and is comparable or superior to vanilla fine-tuning.
Diverse reasoning improves performance even further, often exhibiting significant gains.
We report our full findings on open-source models in Appendix~\ref{open_source_experiments}.

\subsection{Analysis}
\label{analysis}

\paragraph{Sample study}
To identify the strengths and weaknesses of our method, we perform a thorough sample study across all datasets and methods.
Across all arithmetic tasks, we find that a large portion of errors arises from calculations.
MultiArith and SVAMP also show many semantic errors, but these are significantly reduced with diverse reasoning.
For difficult tasks such as GSM8K and AQUA, we found that all methods tend to struggle. 
We found that our method is highly effective in text-based tasks, excluding commonsense reasoning, as well as tasks that contain common linguistic patterns.
On the other hand, we find that students under Zero-shot-CoT often repeat questions or produce incoherent repetitive statements.
While Few-shot-CoT elicits step-by-step sentences, the student models rarely seem to understand the semantics of the question, and generations often contain logical or commonsense errors.
For details on our sample study, see Appendix~\ref{sample_study}.


\paragraph{Nuances of fine-tuning on CoT reasoning}
We shed light on nuances that have often been overlooked in previous or concurrent work \citep{wei2022chain, li2022explanations, magister2022teaching}.
First, we acknowledge the possibility that \textit{correct} samples may contain incorrect reasoning.
In fact, we find that 27.6\% of \textit{correct} teacher completions for Date Understanding contained reasoning errors.
However, ablations on rationale filtering suggest that these incorrect rationales can aid in student supervision (Appendix \ref{rationale_filtering}).
Secondly, we find that common maximum sequence lengths used for CoT generations often lead to incomplete answers.
We observe that reasoning length differs among datasets, and longer generations typically improve accuracy, but may not be beneficial for fine-tuning (Appendix \ref{generation_length}).
Lastly, we find that many datasets are comprised of samples that share common templates, potentially compromising the validity of our random train-test splits.
To address this, we evaluate our method on manual template-wise data splits, and confirm that students retain meaningful reasoning capabilities (Appendix \ref{templated_datasets}).

%% file: section5_discussion.tex
\section{Discussion}
\label{discussion}

\subsection{Accessibility of Fine-tune-CoT}

Owing to the versatility of the teacher generation method, i.e., Zero-shot-CoT, our method can be readily applied to any complex task without task-specific engineering.
Rationales can be readily generated using publicly available APIs such as those provided by OpenAI or Anthropic.
This makes it viable to obtain CoT training data in low-resource scenarios, which not only outperforms standard fine-tuning, but elicits the student to output interpretable explanations.
Fine-tuning and inference on student models can also be performed on much more accessible hardware, in contrast to very large models.
This can reduce long-term inference costs and minimize environmental impact while making our method fully accessible to a wide community.

\subsection{Viability of Fine-tune-CoT}

While Fine-tune-CoT elicits notable complex reasoning capabilities in small models, performance on some difficult datasets would not be considered viable for real-world use, such as 30.33\% on SVAMP.
However, our findings in Section~\ref{results} indicates significant potential for improvement, as our method is shown to be uniquely scalable with (1) diverse reasoning, (2) dataset size, (3) teacher model performance, and (4) student model size.
The use of diverse reasoning and better teacher models is especially promising, as these can benefit from improved teacher LLM performance and inference costs in the future.
In addition, it is possible to incorporate recent CoT methods, which lead to significant performance improvements, in student models, which we discuss in Section \ref{cot_methods}.

\subsection{Tradeoffs of Fine-tune-CoT}
\label{tradeoffs}

The aforementioned opportunities to enhance Fine-tune-CoT also pose many important tradeoffs. We leave further analysis to future work.

\paragraph{Degree of diverse reasoning}
The performance benefits of diverse reasoning come at the cost of additional teacher inference.
Therefore, diverse reasoning poses a tradeoff between development cost vs inference cost/quality. In other words, performance gains from diverse reasoning may be utilized to enhance student performance or alleviate the need for larger student models.
This must also be taken into account for fair evaluation of similar distillation methods in the future.

\paragraph{Data acquisition}
Data annotation and diverse reasoning can both be used to enlarge fine-tuning data, but each have their associated costs.
We note that the cost of diverse reasoning is linear to the number of generated rationale \textit{and} the number of original samples.
Despite this, it can still be a cost-effective alternative to hand-annotating additional data.
A preliminary cost analysis in Appendix~\ref{data_annotation_vs_diverse_reasoning} shows that the pareto front of data-acquisition-cost to performance always incorporates diverse reasoning.
We expect that the cost benefits of diverse reasoning will continue to improve with improvements in teacher model performance and efficiency. 


\subsection{Emergence of CoT reasoning}
\label{emergence}

The emergence of abilities such as CoT reasoning has become a point of interest in recent works \cite{wei2022chain, wei2022emergent, schaeffer2023emergent}.
We note that the efficacy of Fine-tune-CoT on small models does not disprove this emergence, as our method is based on fine-tuning. However, 
we believe our results can provide some insight into this phenomena.

\paragraph{Why does Fine-tune-CoT work in small models?}
In a seminal work, \citet{wei2022chain} suggests that CoT reasoning is an emergent ability of scale---more specifically, a complicated phenomena involving a variety of emergent abilities, such as semantic understanding, symbol mapping, arithmetic ability.
However, our sample studies suggest that Fine-tune-CoT elicits these \textit{emergent} abilities even in relatively small models (see Appendix \ref{sample_study}).
We explain this from two perspectives.
First, \citet{wei2022chain} demonstrated the emergence of reasoning abilties by identifying a reduction in the frequency of reasoning errors with larger model scale.
Similarly, we find that more potent forms of supervision also lead to a \textit{gradual} reduction in reasoning errors.
For example, we found a clear distinction between Zero-, Few-shot-CoT and Fine-tune-CoT (with diverse reasoning) in the frequency and severity of semantic errors, i.e., understanding complex questions, and calculation errors.
This suggests that explicit supervision on reasoning can also lead to the emergence of reasoning abilities.
Second, we qualitatively find that students show capabilities that are reminiscent of the larger teacher model.
We found that students can recognize common semantics and reasoning cues of the given task, and is able to imitate the process of splitting large tasks into subtasks.
This suggests that it is possible to learn reasoning abilities pertaining to a particular domain.
We posit that this is possible in small models due to the limited domain of reasoning, and may not be applicable in reasoning tasks that require large domains of knowledge.


\paragraph{Distillation of emergent abilities}
Chain-of-thought reasoning has been recognized as a prime example of emergent abilities in very large language models~\cite{wei2022emergent}.
Our findings show that it is possible to distill this ability, under certain domains, to much smaller models simply through fine-tuning.
The potential for distillation implies that future advancements in language models may lead to emergent abilities that are not only pertinent to those larger models, but could also have a broader impact, cascading benefits to smaller models.



%% file: section6_conclusion.tex
\section{Conclusion}
\label{conclusion}

We have proposed Fine-tune-CoT, a method that uses LLMs as \textit{reasoning teachers} to transfer the broad reasoning capabilities previously found in $>$100B models to student models as small as 0.3B.
We propose diverse reasoning as a novel approach to maximize these teaching effects, exploiting the unique characteristics of this new learning setup to \textit{vastly} improve performance.
Our extensive experiments show that Fine-tune-CoT elicits significant reasoning performance in small models, thus demonstrating the distillation of CoT reasoning which has been considered an \textit{emergent} ability of scale.
By leveraging publicly available models with zero-shot prompting, we demonstrate a task-agnostic approach to elicit reasoning performance in small models, making complex reasoning feasible for real-world deployment and accessible to the broader community.

%% file: section7_limitations.tex
\section{Limitations}
\label{limitations}

\subsection{Towards concise answers}
\label{concise_answers}

Sample studies show that rationales output from student models may occasionally be repetitive and digressive.
This is undesirable in terms of inference-time efficiency as well as interpretability.
As a minor optimization to inference computation, we construct our fine-tuning sample templates using special-character based delimiters instead of natural language used in concurrent work \citep{huang2022large} to minimize sequence length.
Preliminary findings showed this had no significant impact on reasoning performance.
More importantly, it is desirable to train student models to generate concise answers in terms of substance.
Appendix~\ref{generation_length} hints at the possibility for this, showing that fine-tuning on shorter reasoning samples causes the student model to also produce shorter rationales.


\subsection{Exploring a wider array of models}
We note that the performance of our method is currently not state-of-the-art. However, it can benefit from advances in teacher models as well as other prompting methods. For example, future work should include a wider array of teachers, such as the highly versatile ChatGPT, which typically generates detailed long responses that may be able to impart more knowledge to the student. More recent models such as GPT-4 have demonstrated significant advances in complex reasoning abilities, which may improve the efficacy of Fine-tune-CoT on very difficult datasets, such as GSM8K. Conversely, our method could prove even more advantageous when applied to recent models with improved efficiency, such as those based on the recent LLaMA model~\citep{touvron2023llama}, which has sparked a proliferation of work focused on compact language models.
Both of these avenues are promising for future work.

\subsection{Better CoT inference methods}
\label{cot_methods}
The use of diverse reasoning and better teacher or student models is especially promising, as it is possible to leverage future improvements in model performance and decreased inference costs. However, we can also consider other ways to boost performance, such as using different prompting methods. For example, previous work shows that Few-shot-CoT \citep{wei2022chain} can improve accuracy over Zero-shot-CoT by a wide margin, e.g., going from 78.7\% to 93.0\% on MultiArith \citep{kojima2022large}. However, our choice to use Zero-shot-CoT to generate reasoning samples from the teacher model is motivated by the fact that Few-shot-CoT requires a significantly larger inference context. With the current pricing models based on token usage, the typical setup of 8-shot CoT would cost approximately 8 times more
compared to Zero-shot-CoT. Therefore, we see a tradeoff between using the inference
budget for Few-shot-CoT and using it for diverse reasoning with Zero-shot-CoT.
On the other hand, we also note that recent works introduce various ways to improve CoT reasoning performance substantially (often to near-perfect levels), which can be applied to our student models. These include refinement over repeated inferences \citep{wang2022self, li2022diverse} and self-improvement \citep{zelikman2022star, huang2022large}. In particular, self-consistency~\citep{wang2022self} can be utilizied on unlabeled samples to maximize the teaching signal.
In contrast, we aim to achieve CoT reasoning without the inference time cost incurred by
very large LMs. 
Future work is needed to incorporate these methods into Fine-tune-CoT while minimizing development and inference costs.

\subsection{Connection with knowledge distillation}
We assume that there is a lot of potential in strengthening the connections between knowledge distillation and our method. We have already seen in this work that our method shares some characteristics with KD, such as the fact that the knowledge of intermediate reasoning imparted by using also incorrect samples can have positive effects on student accuracy, akin to ``dark knowledge''~\citep{menon2021statkd} that is transferred by training on teacher output logits and not one-hot labels.  We have seen that this leads to a quantity-quality tradeoff when it comes to the ability of the student model to generalize: having fewer but perfectly curated reasoning samples is not necessarily as helpful as having a larger amount of reasoning samples that might not always be fully correct. On the other hand, we have also found that more accurate teachers do lead to more accurate students, which is not always the case in KD~\cite{muller2019label}. It would therefore be of interest for future work to formalize the connection of Fine-tune-CoT with classic KD methods, and potentially test the use of a different distillation loss function that takes the teacher's actual output into account.



\section{Ethics Statement}
\label{ethics}

Our work presents various challenges and opportunities in terms of bias and toxicity in language models.
It is widely known that LLMs trained on large corpora have been shown to capture biases found in the training data \citep{gpt3__brown2020language, chowdhery2022palm}.
Since our student models are trained on reasoning samples generated by these LLMs, it is possible that such characteristics of the teacher model can get passed along to the student.
This is an important point to consider when selecting the teacher model for our method.

Our training setup, however, does offer a unique opportunity to minimize bias and toxicity in student models, by influencing the samples used for fine-tuning.
One approach would be to augment the curating step of Fine-tune-CoT to filter out biased or toxic samples. It is possible to automate this via neural network-based verifiers, previously used to filter correct output \citep{gsm8k__cobbe2021training, li2022diverse}.
Alternatively, one may consider optimizing the CoT prompts to minimize bias and toxicity in teacher-generated rationales.

We note that bad actors can also potentially take advantage of our method to utilize complex reasoning for malicious purposes and deploy it at scale, using small models.
This highlights the importance of safeguarding the potential capabilities of LLMs by major providers.
To prevent the distillation of malicious reasoning abilities in small (or large) students, future work in identifying usage patterns involved in these distillation schemes may help providers apply more stringent precautions to these use cases.

%% file: appendix.tex
\section{Experimental Details}
\label{experimental_details}

\subsection{Generation}

\paragraph{Maximum sequence length} For the maximum sequence length of teacher-generated rationales, $\hat{r}_i$, we use $L_r = 128$, following \citet{kojima2022large}, unless stated otherwise.
For the maximum sequence length of the student model predictions, we use $L_p = 1024$, unless stated otherwise.
We retroactively applied $L_p = 1024$ as the default, after discovering that $L_p = 128$ is insufficient for many tasks, as discussed in Appendix \ref{generation_length}.

\paragraph{Sampling temperature} We apply greedy decoding for all generations, except diverse reasoning, to obtain deterministic results following \citep{wei2022chain, kojima2022large}.
For diverse reasoning, we use temperature sampling with $T = 0.7$ to obtain diverse samples, following a similar approach from \citet{wang2022self}.

\subsection{Answer cleansing}

We follow the method used in \citet{kojima2022large} to cleanse answers generated by models to assess their correctness.

\subsection{Few-shot-CoT exemplars}
\label{few_shot_cot}

For Few-shot-CoT prompting, we use exemplars provided by \citet{wei2022chain}, with some minor formatting adaptations for consistency with our other experiments. For Last Letter Concatenation and Coin Flip, for which Few-shot-CoT prompts are not provided, we use 8 training samples used in our 8-shot data experiments shown in Figure \ref{fig:dataset_size} and adapt them for Few-shot-CoT using the format of \citet{wei2022chain}. This was not applicable to Tracking Shuffled Objects, therefore it was omitted from Few-shot-CoT experiments.

\subsection{Fine-tuning OpenAI models}
\label{fine_tuning_openai}

We use default hyperparameters set by the OpenAI API for both vanilla fine-tuning and Fine-tune-CoT. While the specifics of the fine-tuning API is not publicly known, some details on hyperparameters are documented in the API reference \footnote{\href{https://platform.openai.com/docs/api-reference/fine-tunes/create}{https://platform.openai.com/docs/api-reference/fine-tunes/create}}.
According to the default settings, our models are trained for 4 epochs.
The batch size and learning rate determined based on the number of examples used for training.
The batch size is set to 0.2\% of the number of training examples capped at 256.
The learning rate is set to 0.05, 0.1, or 0.2 times that of the learning rate used to pre-train the base model, depending on the batch size.
Training loss is also applied to the prompt portion of the training examples, i.e., the question, with a small weight of 0.01.
Based on API pricing , we posit that OpenAI employs a form of parameter efficient fine-tuning such as LoRA \citep{hu2021lora} for their fine-tuning API instead of updating all model parameters.

\subsection{Fine-tuning open source models}
\label{fine_tuning_open_source}

For vanilla fine-tuning and Fine-tune-CoT on open source models, we strictly control for hyperparameters.
Across all experiments, we fine-tune the entire model with a fixed learning rate of 3e-4 and batch size of 8.
Upon inspection of model performance under various learning rates and batch sizes, we found that optimal parameters varies among datasets, even between those with similar number of reasoning samples.
We train all models for a maximum of 20 epochs, which we found to be sufficient for test accuracy to plateau.
We report the best test accuracy from 20 epochs, but found that performance varies significantly between epochs.
Overall, we found that performance by epoch is stable for larger models, and that instruction-tuned Flan-T5 is more stable compared to T5.
Similar to learning rate and batch size, the optimal number of epochs also varies between datasets, even those with similar number of reasoning samples.
Based on the above, we note that our reported performances of fine-tuned open-source models may be significantly under-estimated compared to those with optimal hyperparameters, and recommend practitioners to optimize hyperparameters using a separate validation set, per each training setting.

\begin{table*}[ht]
\resizebox{\linewidth}{!}{
\begin{tabular}{lccclll}
\toprule
Dataset & Choices & Training Samples & Test Samples & Data Split & License & References \\
\midrule
SingleEq & - & 356 & 152 & 70:30 & None & \citet{singleeq__koncel2015parsing} \\
AddSub & - & 276 & 119 & 70:30 & Unspecified & \citet{addsub__hosseini2014learning} \\
MultiArith & - & 420 & 180 & 70:30 & Unspecified & \citet{multiarith__roy2016solving} \\
GSM8K & - & 7473 & 1319 & Original & MIT & \citet{gsm8k__cobbe2021training} \\
AQUA-RAT & 5 & 10000 & 254 & Custom & Apache-2.0 &  \citet{aquarat__ling2017program} \\
SVAMP & - & 700 & 300 & 70:30 & MIT & \citet{svamp__patel2021nlp} \\
Date Understanding & 5--6 & 258 & 111 & 70:30 & Apache-2.0 & \citet{big__srivastava2022beyond} \\
Tracking Shuffled Objects & 3 & 525 & 225 & 70:30 & Apache-2.0 & \citet{big__srivastava2022beyond} \\
Last Letter Concatenation & - & 350 & 150 & 70:30 & Unspecified & \citet{wei2022chain, kojima2022large} \\
Coin Flip & 2 & 350 & 150 & 70:30 & Unspecified & \citet{wei2022chain, kojima2022large} \\
CommonSenseQA & 5 & 9741 & 1221 & Original & Unspecified & \citet{talmor2018commonsenseqa} \\
StrategyQA & 2 & 1603 & 687 & 70:30 & Apache2.0 & \citet{strategyqa__geva2021did} \\
\bottomrule
\end{tabular}
}
\caption{Description of datasets used in our study.}
\label{tab:datasets}
\end{table*}

\section{Datasets}
\label{datasets}

We provide a summary of datasets used in our experiments, including their original licenses, in Appendix Table~\ref{tab:datasets}. We consider the 12 datasets used in \citet{kojima2022large} to measure reasoning performance. For Last Letter Concatenation and Coin Flip, we use the publicly available data provided by \citet{kojima2022large}.

\paragraph{Train-test split} Contrary to previous works on prompt-based CoT such as \citet{wei2022chain, kojima2022large}, our fine-tuning approach requires distinct sets of samples for training and testing.
If separate subsets for training and testing (or development) are provided, we use those.
Otherwise, we perform a samplewise random split with a train-test ratio of 70:30.
For AQUA, due to the disproportionately large size of the original training set, we randomly sample 10,000 instances for training in our experiments.
Note that due to the highly templated nature of many datasets, this naive data split may not be appropriate for evaluating reasoning capabilities.
This is an important nuance of fine-tuning on CoT reasoning, which we address in Appendix \ref{templated_datasets}.

\begin{table}[ht]
\resizebox{\linewidth}{!}{
\begin{tabular}{llll}
\toprule
Model Family & Params & Role & Variant / Name in API \\
\midrule
GPT-3 & 175B & Teacher & \texttt{davinci} \\
InstructGPT & 175B & Teacher & \texttt{text-davinci-001} \\
InstructGPT & 175B & Teacher & \texttt{text-davinci-002} \\
InstructGPT & 175B & Teacher & \texttt{text-davinci-003} \\
\midrule
GPT-3 & 6.7B & Student & \texttt{curie} \\
GPT-3 & 1.3B & Student & \texttt{babbage} \\ 
GPT-3 & 0.3B & Student & \texttt{ada} \\
\midrule
T5 & 60M & Student & Small \\
T5 & 220M & Student & Base \\ 
T5 & 700M & Student & Large \\
\midrule
Flan-T5 & 60M & Student & Small \\
Flan-T5 & 220M & Student & Base \\ 
Flan-T5 & 700M & Student & Large \\
\midrule
GPT-2 & 124M & Student & (Small) \\
GPT-2 & 355M & Student & Medium \\ 
GPT-2 & 774M & Student & Large \\
\bottomrule
\end{tabular}
}
\caption{Description of models used in our study.}
\label{tab:models}
\end{table}

\section{Models and API Usage}
\label{models}

Appendix Table~\ref{tab:models} describes all teacher and student models used in our study.
We use InstructGPT \citep{ouyang2022instruct} as the default teacher model in our experiments, due to its superior zero-shot reasoning performance, compared to GPT-3 \citep{gpt3__brown2020language} of the same size \citep{kojima2022large}.
Specifically, we use \texttt{text-davinci-002} at the default, as it was the best available model at the start of our experiments.
We were unable to consider small InstructGPT models for fine-tuning, as it is not offered by the OpenAI API.
We attach model size information based on (\href{https://blog.eleuther.ai/gpt3-model-sizes/}{https://blog.eleuther.ai/gpt3-model-sizes/}), following \citet{kojima2022large}.

Our total expenditure for API usage, including all preliminary experiments, was \$1,981 USD.
The majority of this expenditure occurred after September 1st, 2022, from which the pricing for inference on \texttt{davinci} models was \$0.02/1K tokens, among others.
Between teacher model inference, student model \{fine-tuning, inference\}, the majority of API usage in terms of cost was focused on teacher model inference.


\section{Sample Study}
\label{sample_study}
To understand where our method makes mistakes, where diverse reasoning can improve performance, and where our method always performs well, we observe randomly sampled instances and analyze the reasoning performance on them. To do so, we compare its generations for these samples with (1) the output of the large teacher model, (2) a student model using Zero-shot-CoT  (3) a student model using Few-shot-CoT, and (4) a student model using fine-tuning without chain of thought reasoning. Our analysis reflects our overall findings, which we exemplify with representative examples in Tables~\ref{table:ex1}--\ref{table:ex4}. 

\subsection{Error analysis}
For our analysis of the most common types of errors, we take a look at datasets where we find particularly bad performance of our vanilla method, also in comparison to other students. We also discuss the benefits of using diverse reasoning in~\ref{improve_diverse}. We summarize our observations below. 

\paragraph{Difficult datasets}
First, we observe that the sets GSM8K and AQUA appear to be too difficult for any small student model, in particular, given that the teacher model gets below 50\% accuracy on both. In fact, even correct answers are usually correct only by chance, due to the high complexity of the tasks (Appendix Tables \ref{table:ex1}a,b). For AQUA in particular, we note that while we occasionally find meaningful reasoning in the 6.7B student model, students clearly cannot sufficiently learn to solve the tasks. We do note however that of all the student methods, Fine-tune-CoT still gets the best performance in these two datasets. A similar, if less salient, issue arises for StrategyQA. Here, the teacher also performs only 3\% above the random guess accuracy of 50\%. The smaller student models actually manage to improve on this performance as long as they do not use Zero-shot-CoT, in particular vanilla fine-tuning, but the errors arising in Fine-tune-CoT often look very similar to the ones in the large teacher model. We see that all models usually merely retrieve information related to the question, but cannot synthesize an answer from it (Appendix Tables \ref{table:ex1}c,\ref{table:ex2}a). 

\paragraph{Arithmetic mistakes}
Next, we note that small models overall exhibit weak arithmetic skills. This has already been discussed in previous literature, where calculation capability has been found to scale with model size~\cite{wei2022emergent}. Especially in SingleEq (Appendix Table \ref{table:ex2}b) and AddSub (Appendix Table \ref{table:ex2}c), a majority of errors in the output of student models using Fine-tune-CoT simply arise from wrong calculations, less so bad reasoning. This is also a major factor in the bad performance our method exhibits on SVAMP as well as GSM8K; even correct multistep reasoning cannot compensate for the fact that the model's arithmetic tends to be wrong on intermediate steps (Appendix Tables \ref{table:ex2}d, \ref{table:ex3}a). Only the teacher model then does better on these tasks, given its much larger size, even though it does not get perfect accuracy either. However, we note here that very large language models, such as PaLM 540B, can be trained on arithmetic and scientific data to be able to reason correctly about a wide range of mathematical tasks in a step-by-step fashion 
~\citep{lewkowycz2022minerva}. 

\paragraph{Problematic benchmarks, impact of commonsense reasoning errors} Meanwhile, when looking at our method's performance in CommonsenseQA, we note that producing consistent multi-step reasoning is not always the issue. We find that the student model utilizing Fine-tune-CoT can often generate logical reasoning paths for many of the samples that are marked as false (Appendix Table~\ref{table:ex4}b). Rather, the exact answer is often subjective, making it difficult to guess the correct output from logical reasoning alone (Appendix Table~\ref{table:ex4}c).
CommonsenseQA thus is not always an ideal benchmark when judged on accuracy, but gives insight into how well the model can produce reasoning. We also note a difference compared to Few-shot-CoT in terms of the impact of reasoning errors: the latter only performs around 5\% above random, lacks understanding of the question in many cases, and makes more severe logical and commonsense mistakes compared to our method.
In fact, Fine-tune-CoT comes close to the teacher due to the relatively lower impact of errors that do arise (Appendix Table~\ref{table:ex4}d).
This suggests that Fine-tune-CoT enables stronger task-solving capabilities and avoids making serious commonsense mistakes that prevent it from arriving at a reasonable conclusion.

\paragraph{Aligned failures} Importantly, we note that for each dataset, there seems to be a difference between ``easy'' and ``hard'' instances.
When we consider the accuracy of the teacher and other student models (using fine-tuning, Zero-shot- or Few-shot-CoT) on tasks where our method fails, we find that it is always lower than on tasks where our method is successful.
That is, successes and failures tend to be aligned across the different methods. We can hypothesize that factors such as content bias may play a role here; language models have been found to fail depending on context and content of the task, in a way similar to human reasoners~\citep{dasgupta2022human}. We can identify samples that hint at this issue when we look at questions that include phrasing that seems contradictory or counterintuitive to the context that the model expects (see Appendix Table~\ref{table:ex4}d, where the number of movies watched is larger than the number of available movies). Additionally, previous work shows that GPT-3 exhibits a performance gap between instances including terms that are frequent in the pretraining corpus, and instances including less frequent terms~\citep{raszeghi2022impact}. This can contribute to uneven performance on a multitude of (especially numerical) tasks across different methods and model sizes. 
We can then surmise the observed absolute differences in accuracy to stem from the various sources of errors for each method. For example, fine-tuning has much less room for error than Fine-tune-CoT, which can additionally make mistakes on intermediate reasonings such that errors compound. 

\subsection{Improvements from diverse reasoning}
\label{improve_diverse}
\paragraph{Semantic issues} We find that models seem sensitive to how a question is formulated. This is noticeable in all datasets, in particular in SVAMP and to a certain degree in MultiArith. Besides arithmetic mistakes, we observe that such semantic issues are one of the main factors for uneven performance of vanilla Fine-tune-CoT on these two datasets.

In particular, we observe this issue when there is redundant information present in the question (Appendix Table~\ref{table:ex3}b). Such cases elicit wrong reasoning, or lead the model to become stuck on the question, similarly to what usually happens with Zero-shot-CoT in the student model (i.e. repeating the question, or coming up with information that only vaguely pertains to the question). Other common sources of errors are when hidden variables make up the first part of the task (i.e. those tasks that force the model to calculate a previously unknown value that is described in the first sentence (Appendix Table~\ref{table:ex3}c), or when the model encounters overloaded words (e.g., ``landing'' in Appendix Table~\ref{table:ex3}d). We also observe samples where the model gets stuck on an intermediate result (Appendix Table~\ref{table:ex4}a). This observation agrees with previous findings that language models have recency bias~\cite{zhao2021calibrate}. 

However, this source of errors can be compensated for by using diverse reasoning. When comparing the generations from Few-shot-CoT, vanilla Fine-tune-CoT and Fine-tune-CoT with diverse reasoning on MultiArith, we find that diverse reasoning enables the model to understand the question better.
While calculation errors are still relatively frequent, the generations show clear advantages in terms of semantic understanding and being able to reason logically as a consequence.
This is especially clear when compared to Few-shot-CoT, which exhibits problems both in understanding the question and formulating coherent expressions, especially when three or more terms are involved in the calculation, as mentioned in \citet{kojima2022large}. By contrast, Fine-tune-CoT with diverse reasoning makes for a significantly smoother reasoning performance than using Few-shot-CoT or even vanilla Fine-tune-CoT. This results in vastly improved accuracy on both MultiArith and SVAMP.

\subsection{Strengths}
 Having analyzed the main sources of errors, we can now focus on the datasets that elicit good performance from our method, regardless of whether we use diverse reasoning. 
 
 \paragraph{Text-based datasets} As arithmetic errors are one of the main reasons for the decrease in performance of small student models, it comes as little surprise that our vanilla method without diverse reasoning performs well on datasets that are mainly text-based and do not require actual calculation skills. This includes Date Understanding (60.4\%) (Appendix Table~\ref{table:ex5}a), Last Letter Concatenation (52.67\%) (Appendix Table~\ref{table:ex5}b), Coin Flip (98.7\%) (Appendix Table~\ref{table:ex5}c), and Shuffled Objects (64.4\%)  (Appendix Table~\ref{table:ex5}d).
 Our methods performs significantly above random choice on these sets, and additionally beats the teacher on Shuffled Objects and Coin Flip.
 We find that accuracy metrics for these sets are mostly faithful: while the elicited reasoning is not always very detailed, and occasionally misses some reasoning steps (Appendix Table~\ref{table:ex5}e) , the model draws correct conclusions from mostly correct steps. We also note that similar to MultiArith and SVAMP, performance on these four datasets can be even further boosted with diverse reasoning, outperforming the teacher model across all four. 
 
 \paragraph{Patterns}
 These datasets also have very clear patterns in their tasks, which helps Fine-tune-CoT to perform well by providing cues on how to solve a specific task.
 We note that in contrast, classic fine-tuning does not have an advantage in these datasets, and it gets significantly lower accuracy than Fine-tune-CoT on all four. The same is also true for MultiArith, which we have used as a benchmark in the main text. While arithmetic errors cause the absolute accuracy of our method to be lower than the teacher, it significantly outperforms fine-tuning on MultiArith even without using diverse reasoning. Indeed, we find that also in the presence of arithmetic errors, our model reasons correctly in many cases. We can surmise that the patterned nature of the tasks in MultiArith helps the student model to understand what is asked of it, eliciting the correct reasoning. 
Additionally, we note that the presence of such patterns in successful datasets does not mean that our method overfits to existing templates.
In our template-split analysis (Appendix~\ref{templated_datasets}), we in fact show that while tasks look similar to one another in certain datasets such as Date Understanding, the student model's reasoning does not rely on simply matching templates or memorizing particular solutions.
This implies that our method can generalize to previously unseen tasks; the patterns in the datasets do not produce overfitting, but can be surmised to act as cues for the model's understanding of its current task.
Thus, we observe that the reasoning skills of a student using Fine-tune-CoT can overcome the smaller model capacity (which proves to be completely prohibitive, e.g., for Zero-shot-CoT to have any success on the various tasks).

\begin{table*}
\begin{minipage}[b]{1.0\linewidth}
\centering
\resizebox{\linewidth}{!}{

\begin{tabular}{llcccccccccccc}
\toprule
Model & Max & Single & Add & Multi & \multirow[c]{2}{*}{GSM8K} & \multirow[c]{2}{*}{Aqua} & \multirow[c]{2}{*}{SVAMP} & Date & Shuffled & Last & Coin & Common & Strategy \\
Params & Tokens & Eq & Sub & Arith &  &  &  & Understanding & Objects & Letter & Flip & SenseQA & QA  \\
\midrule
\multicolumn{14}{c}{\textbf{Teacher: InstructGPT (\texttt{text-davinci-002})}} \\
\midrule
\multirow[c]{4}{*}{175B} & \multirow[c]{2}{*}{128} & 81.18 & 75.72 & 76.90 & 42.42 & 29.63 & 64.00 & 65.89 & 54.10 & 57.43 & 89.71 & 59.86 & 53.40 \\
 &  & \textbf{\small{(84.83)}} & \small{(90.22)} & \small{(95.24)} & \textbf{\small{(69.85)}} & \textbf{\small{(44.04)}} & \textbf{\small{(86.57)}} & \small{(98.06)} & \small{(97.14)} & \small{(99.71)} & \small{(97.14)} & \textbf{\small{(82.55)}} & \textbf{\small{(71.55)}} \\
 & \multirow[c]{2}{*}{2048} & 81.18 & 75.72 & 76.48 & 47.73 & 34.77 & 66.00 & 63.28 & 54.10 & 57.43 & 89.71 & 59.40 & 53.03 \\
 &  & \textbf{\small{(84.83)}} & \small{(90.22)} & \small{(94.29)} & \small{(99.34)} & \small{(96.42)} & \small{(99.00)} & \small{(97.14)} & \small{(97.14)} & \small{(99.71)} & \small{(97.14)} & \small{(99.92)} & \small{(99.69)} \\
\midrule
\multicolumn{14}{c}{\textbf{Student: GPT-3 (\texttt{ada}, \texttt{babbage}, \texttt{curie})}} \\
\midrule
\multirow[c]{4}{*}{0.3B} & \multirow[c]{2}{*}{128} & 7.24 & 6.72 & 5.56 & 3.11 & 16.54 & 4.33 & 17.12 & 48.89 & 50.67 & 99.33 & 30.30 & 47.16 \\
 &  & \small{(96.05)} & \small{(99.16)} & \small{(96.11)} & \textbf{\small{(74.75)}} & \textbf{\small{(45.67)}} & \small{(91.33)} & \small{(100.00)} & \small{(100.00)} & \small{(100.00)} & \small{(100.00)} & \textbf{\small{(86.73)}} & \textbf{\small{(87.63)}} \\
 & \multirow[c]{2}{*}{1024} & 7.24 & 6.72 & 6.11 & 3.11 & 23.62 & 5.00 & 17.12 & 49.33 & 50.67 & 99.33 & 32.68 & 52.55 \\
 &  & \small{(98.68)} & \small{(99.16)} & \small{(97.22)} & \small{(99.77)} & \small{(100.00)} & \small{(97.33)} & \small{(100.00)} & \small{(100.00)} & \small{(100.00)} & \small{(100.00)} & \small{(100.00)} & \small{(99.71)} \\
\midrule
\multirow[c]{4}{*}{1.3B} & \multirow[c]{2}{*}{128} & 11.18 & 11.76 & 13.89 & 4.02 & 15.35 & 7.33 & 38.74 & 53.78 & 50.67 & 100.00 & 40.95 & 47.02 \\
 &  & \small{(92.76)} & \small{(96.64)} & \small{(98.89)} & \textbf{\small{(75.36)}} & \textbf{\small{(48.03)}} & \small{(90.33)} & \small{(100.00)} & \small{(99.56)} & \small{(100.00)} & \small{(100.00)} & \textbf{\small{(86.57)}} & \textbf{\small{(83.99)}} \\
 & \multirow[c]{2}{*}{1024} & 11.18 & 11.76 & 13.33 & 4.70 & 19.69 & 8.00 & 38.74 & 52.44 & 50.67 & 100.00 & 43.08 & 52.69 \\
 &  & \small{(98.68)} & \small{(98.32)} & \small{(99.44)} & \small{(99.92)} & \small{(99.61)} & \small{(99.00)} & \small{(100.00)} & \small{(100.00)} & \small{(100.00)} & \small{(100.00)} & \small{(99.92)} & \small{(98.98)} \\
\midrule
\multirow[c]{4}{*}{6.7B}  & \multirow[c]{2}{*}{128} & 21.05 & 20.17 & 34.44 & 7.20 & 16.93 & 12.67 & 60.36 & 64.00 & 52.00 & 98.00 & 51.27 & 47.16 \\
 &  & \small{(92.76)} & \small{(97.48)} & \small{(99.44)} & \textbf{\small{(76.19)}} & \textbf{\small{(55.91)}} & \small{(93.67)} & \small{(99.10)} & \small{(100.00)} & \small{(100.00)} & \small{(100.00)} & \textbf{\small{(85.26)}} & \textbf{\small{(84.28)}} \\
 & \multirow[c]{2}{*}{1024} & 20.39 & 21.01 & 33.33 & 6.75 & 24.02 & 12.67 & 60.36 & 64.44 & 52.67 & 98.67 & 56.76 & 55.02 \\
 &  & \small{(98.68)} & \small{(100.00)} & \small{(100.00)} & \small{(99.92)} & \small{(100.00)} & \small{(99.00)} & \small{(100.00)} & \small{(100.00)} & \small{(100.00)} & \small{(100.00)} & \small{(100.00)} & \small{(99.71)} \\
\midrule
Random &  & 0.00 & 0.00 & 0.00 & 0.00 & 20.00 & 0.00 & 17.12 & 33.33 & 0.00 & 50.00 & 20.00 & 50.00 \\
\bottomrule
\end{tabular}

}
\end{minipage}
\caption{
\textbf{Ablation on maximum sequence length.} Accuracy (\%) of Zero-shot-CoT on the teacher model and Fine-tune-CoT on GPT-3 student models, based on maximum sequence length. Values in parentheses refer to the percentage of generated rationales that were completed within the allotted maximum sequence length. Percentages lower than 90\% are marked in bold. Note that the maximum sequence length is applied to the reasoning portion, i.e., step 1, of Zero-shot-CoT and to the entire output of Fine-tune-CoT.
}
\label{tab:inference_length}
\end{table*}

\begin{figure*}[t]
    \centering
    \includegraphics[width=\linewidth]{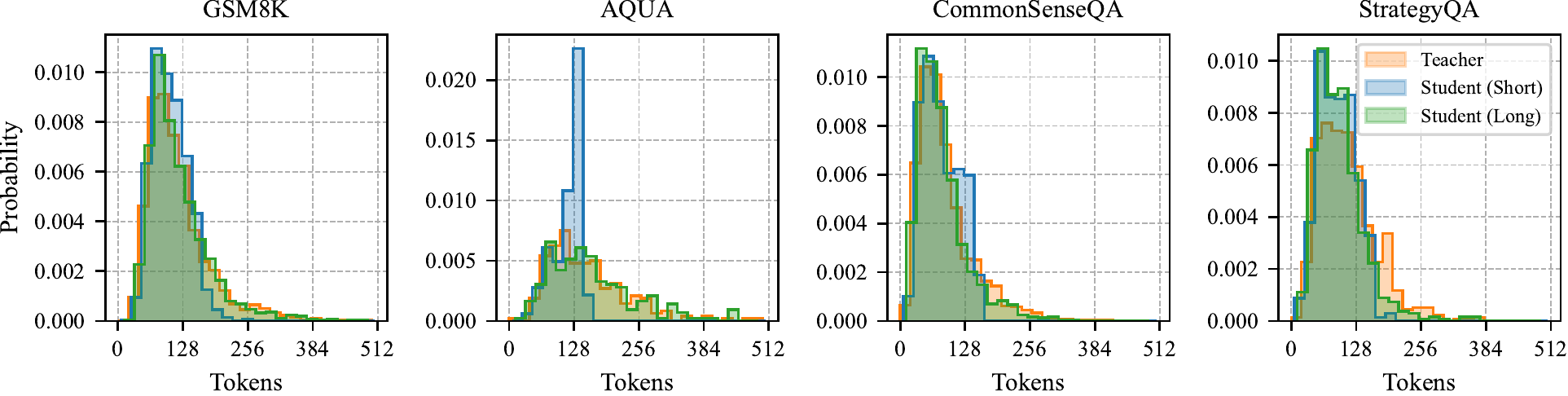}
    \caption{
    \textbf{Effects of teacher reasoning length on student reasoning length.}
    Distribution of the length of generated reasoning sequences from the 175B teacher model and fine-tuned 6.7B student models on four datasets. Student (Short) refers to students that were fine-tuned on reasoning samples with maximum rationale sequence length of $L_r = 128$, and Student (Long) refers to students that were fine-tuned on longer reasoning samples with $L_r = 512$.
    }
    \label{fig:length_distribution}
\end{figure*}


\section{Nuances of Fine-tune-CoT}
\label{nuances}

\subsection{Rationale filtering}
\label{rationale_filtering}

We investigate whether answer-based filtering is sufficient for selecting \textit{good} teacher-generated reasoning samples.
It is possible for the teacher model to answer correctly despite incorrect reasoning, especially in multi-choice questions where the random-guess probability is significant.
To investigate the potential impact of a better filtering scheme (as opposed to our baseline answer-based filtering) we manually annotate the correctness of rationales from the teacher model and evaluate student performance when fine-tuning on \textit{correctly reasoned} samples.
We use the Date Understanding dataset for this ablation, as it is comprised of well-grounded multi-choice questions for which Fine-tune-CoT achieves adequate performance.
Appendix Table~\ref{tab:rationale_filtering} compares the Fine-tune-CoT performance of student models on Date Understanding when using \textit{correct} samples filtered based on answer predictions vs \textit{golden} samples, hand-picked based on the correctness of rationales.
For golden samples, we exclude samples that contain incorrect reasoning steps or irrelevant steps which are misleading.
We find that 28\% of correct samples have incorrect rationales--significantly more than the random-guess performance of 17.12\%, indicating the importance of filtering.
Surprisingly, we however find that answer-based filtering outperforms the more stringent human filtering by 5-11\%, given the same initial samples.
When we match the number of samples post-filtering (via undersampling), we do find that fine-tuning on golden samples outperforms that on correct samples by 5-8\%.
These results suggest that there is a tradeoff between the  quality and quantity of reasoning samples which must be addressed when considering sample-filtering methods.
We also note that this must be considered in tandem with diverse reasoning, which can drastically increase the quantity of reasoning samples.


\begin{table}
\centering
\resizebox{\linewidth}{!}{
\begin{tabular}{llllll}
\toprule
\multirow[c]{2}{*}{Method} & \multirow[c]{2}{*}{Filter}  & \multirow[c]{2}{*}{Samples} & \multicolumn{3}{c}{Model Params} \\
 &  &  & 0.3B & 1.3B & 6.7B \\
\midrule
Zero-shot-CoT &  & 0 & 10.81 & 14.41 & 15.32 \\
\midrule
Fine-tune-CoT & Answer & 170 & 17.12 & 38.74 & 60.36 \\
Fine-tune-CoT & Golden & 123 & 17.12 & 28.83 & 54.95 \\
Fine-tune-CoT & Answer$\textsuperscript{†}$ & 123 & 17.12 & 18.92 & 50.45 \\
Random &  &  & 16.09 &  &  \\
\bottomrule
\end{tabular}
}
\caption{
    \textbf{Effects of rationale filtering.} Accuracy (\%) of GPT-3 student models under Fine-tune-CoT when using samples filtered using answer predictions (Answer), or filtered by humans based on the correctness of the rationale (Golden).
    Answer$\textsuperscript{†}$ refers to using a randomly sampled subset of the correct samples to match the number of golden samples.
}
\label{tab:rationale_filtering}
\end{table}

\subsection{Maximum sequence length}
\label{generation_length}

\begin{table}
\centering
\resizebox{\linewidth}{!}{
\begin{tabular}{llllll}
\toprule
Model & Max & \multirow[c]{2}{*}{GSM8K} & \multirow[c]{2}{*}{AQUA} & Common & Strategy \\
Params & Tokens & & & SenseQA & QA \\
\midrule
\multirow[c]{2}{*}{0.3B} & 128 & 3.11 & 23.62 & 32.68 & 52.55 \\
 & 512 & 3.41 & 15.35 & 32.10 & 52.98 \\
\midrule
\multirow[c]{2}{*}{1.3B} & 128 & 4.70 & 19.69 & 43.08 & 52.69 \\
 & 512 & 3.79 & 18.90 & 43.65 & 53.42 \\
\midrule
\multirow[c]{2}{*}{6.7B} & 128 & 6.75 & 24.02 & 56.76 & 55.02 \\
 & 512 & 7.96 & 18.90 & 58.15 & 54.15 \\
\midrule
Random &  & 1.01 & 20.44 & 20.01 & 50.18 \\
\bottomrule
\end{tabular}
}
\caption{\textbf{Effects of teacher reasoning length on student performance.} Accuracy (\%) of GPT-3 students models under Fine-tune-CoT on four datasets which require longer rationales, when trained on reasoning samples with maximum rationale sequence lengths of $L_r = 128, 512$.}
\label{tab:fine_tune_length}
\end{table}

Following the original setting for Zero-shot-CoT \citep{kojima2022large}, we limit the max sequence length, or max tokens, allowed for the teacher-generated rationale and student reasoning predictions, denoted $L_r$, $L_p$, to 128 initially.
However, we find that this can be insufficient in many datasets.
Allowing for longer inference, we observe that model performance improves significantly on AQUA and commonsense reasoning tasks (Appendix Table~\ref{tab:inference_length}).
Sample inspection shows that rationales with over $\sim$500 tokens are typically repetitive or too digressive.
To investigate the effect of the max length $L_r$ of the teacher rationale on fine-tuning, we compare student performance using $L_r=\{128, 512\}$ (Appendix Table \ref{tab:fine_tune_length}).
The effect of $L_r$ on student performance varies across datasets, and increased $L_r$ does not necessarily improve student performance on tasks that require longer rationales, such as AQUA.
Finally, we examine the length distribution of the generated rationales from the teacher model and student trained on short ($L_r=128$) and long ($L_r=512$) reasoning samples, respectively (Appendix Figure \ref{fig:length_distribution}).
We find that the distribution is different for each dataset.
Notably, we find that while the distributions from the \textit{long} students were similar to that of the teacher, the generated rationale from the \textit{short} students were typically limited to less than $\sim$128 tokens.
These findings are in line with the intuition that different tasks require different lengths of rationales, and suggest that careful consideration is needed in determining parameters related to sequence length.

\begin{figure*}[h]
    \centering
    \includegraphics[width=\linewidth]{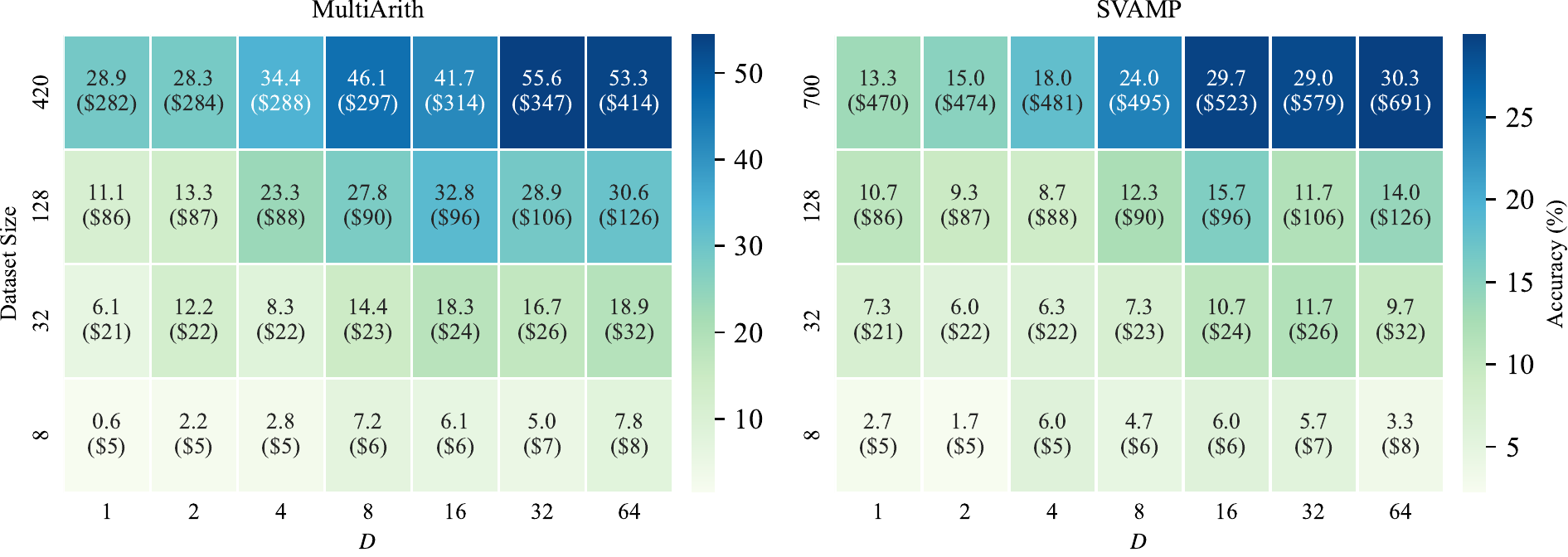}
    \caption{
    \textbf{Cost analysis of data acquisition methods}. Accuracy (\%) of GPT-3 (6.7B) student model under Fine-tune-CoT with varying degrees of diverse reasoning and dataset sizes. Values in parentheses indicate estimated total cost of data acquisition, i.e., data annotation and diverse reasoning inference.
    }
    \label{fig:data_annotation_vs_diverse_reasoning}
\end{figure*}

\begin{table}[h]
\centering
\resizebox{\linewidth}{!}{
\begin{tabular}{llll}
\toprule
Params & Split & MultiArith & Date Understanding \\
\midrule
\multirow[c]{2}{*}{0.3B} & Sample-wise & 5.56 & 17.12 \\
 & Template-wise & 5.35 & 22.22 \\
\midrule
\multirow[c]{2}{*}{1.3B} & Sample-wise & 13.89 & 38.74 \\
 & Template-wise & 7.49 & 35.19 \\
\midrule
\multirow[c]{2}{*}{6.7B} & Sample-wise & 34.44 & 60.36 \\
 & Template-wise & 21.39 & 49.07 \\
\bottomrule
\end{tabular}
}
\caption{
    \textbf{Sample-wise vs template-wise split.} Accuracy (\%) of GPT-3 student models under Fine-tune-CoT on two moderately templated datasets when using a sample-wise vs template-wise train-test split.
}
\label{tab:template_split}
\end{table}

\subsection{Templated datasets}
\label{templated_datasets}
Upon inspection, we found that many datasets contain groups of samples which share common templates.
Therefore, naive samplewise data split has the potential to leak the same templates into the train and test sets, essentially demoting the learning problem into simple pattern matching, rather than complex reasoning.
This brings into question the validity of naive samplewise data split, as it has the potential to leak the same templates into the train and test sets.
To investigate whether the student models are truly learning to reason rather than matching simple patterns, we manually group samples by template and evaluate Fine-tune-CoT using a template-wise data split. 
We consider MultiArith and Date Understanding as they contain a moderate number of templates. Note that all datasets excluding GSM8K, CommonsenseQA, and StrategyQA contain templates to varying degrees.
Appendix Table~\ref{tab:template_split} shows the performance of Fine-tune-CoT when using sample-wise vs template-wise split, using the same train-test ratio of 70:30.
While student performance is typically lower with a template-wise split, it still significantly outperforms random guess performance, as well as prompt-based baselines shown in Appendix Table~\ref{tab:performance}.
This reaffirms that Fine-tune-CoT is able to elicit complex reasoning capabilities in small language models.

\section{Data Annotation vs Diverse Reasoning}
\label{data_annotation_vs_diverse_reasoning}

In Appendix Figure~\ref{fig:data_annotation_vs_diverse_reasoning}, we analyze the cost of data annotation and diverse reasoning, based on current OpenAI API pricing and a low estimate of annotation cost at 30 annotations per hour at an hourly rate of \$20, i.e., \$0.67 per question-answer sample.
When comparing the cost and student performance of models trained with $D=1$ and $D=64$, we can clearly see that using diverse reasoning can enhance the cost-effectiveness of data acquisition.
However, as the cost of diverse reasoning correlates with the size of the dataset, it is important to consider the cost-performance tradeoffs.

\clearpage

\section{Experiments on Open Source Models}
\label{open_source_experiments}

To validate the generality of our method, we apply our method to a wide range of student models beyond variants of GPT-3. While the OpenAI API for GPT-3 inference and fine-tuning is accessible and does not require high-end GPUs, the model weights and implementation are not publicly available and may involve black-box processing.
We therefore conduct experiments from Section~\ref{experiments} on open-source models under a standard setting with fixed hyperparameters, as explained in Appendix~\ref{experimental_details} and report our results in the following.
Tables and figures include results from Section~\ref{experiments} on GPT-3 for reference.

\paragraph{Prompt-based baselines}
\label{performance_open_source_prompt}
A comprehensive performance evaluation of student models across multiple tasks is encapsulated in Table~\ref{tab:performance_all}, comparing Fine-tune-CoT against baseline methods. Performance of standard zero-shot prompting, predominantly insignificant, is omitted when negligible but does exhibit unexpected spikes on Flan-T5, such as 94.22\% on Tracking Shuffled Objects on the smallest model.
Few-shot-CoT likewise demonstrates inconsequential performance across most student models, yet the Flan-T5 models reveal significant performance on some tasks such as 7.51\% on GSM8K and 83.87\% on CommonSenseQA.
This hints at the possibility that instruction tuning may empower models to comprehend and execute CoT prompts, unveiling a latent reasoning capacity within smaller language models.

\paragraph{Fine-tune-CoT vs vanilla fine-tuning}
\label{performance_open_source_fine_tune}
Further examining Table~\ref{tab:performance_all}, we note that vanilla fine-tuning achieves notable performance in encoder-decoder architectures, namely T5 and Flan-T5, achieving more than 80\% on Date Understanding and 100\% on Coin Flip, significantly outperforming vanilla fine-tuning on GPT-2 and GPT-3 student models.
This leads us to believe that the causal attention masking present in decoder-only models could impede complex inter-token reasoning.
CoT reasoning, in this regard, may serve to mitigate this limitation by repeating key information within the decoding context.
Other the other hand, Fine-tune-CoT either surpasses or matches the performance of vanilla fine-tuning across a variety of tasks.
Our method also displays consistent scalability with model size, in contrast to the fluctuating performance between model sizes for baseline methods.
The incorporation of diverse reasoning enhances this scalability.
Particularly, we find that the Flan-T5 models benefit more from Fine-tune-CoT compared to T5 models, implying a favorable role of instruction tuning. When enhanced with diverse reasoning, Fine-tune-CoT excels over vanilla fine-tuning across several complex reasoning tasks, notably observed in the performance of Flan-T5 on Tracking Shuffled Objects (44.00\%$\rightarrow$89.33\%) and GPT-2 on MultiArith (11.67\%$\rightarrow$19.44\%).

\paragraph{Effects of diverse reasoning}
\label{diverse_reasoning_open_source}

\begin{figure}[p]
    \centering
    \begin{subfigure}{0.45\textwidth}
        \includegraphics[width=\textwidth]{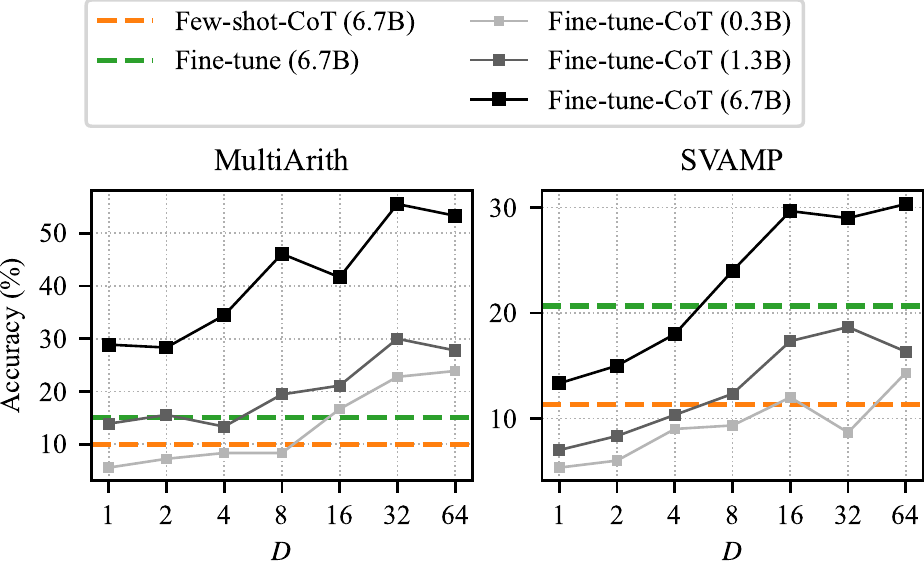}
        \caption{GPT-3}
        \label{fig:diverse_reasoning_gpt3}
    \end{subfigure} \\
    \vspace{0.15cm}
    \begin{subfigure}{0.45\textwidth}
        \includegraphics[width=\textwidth]{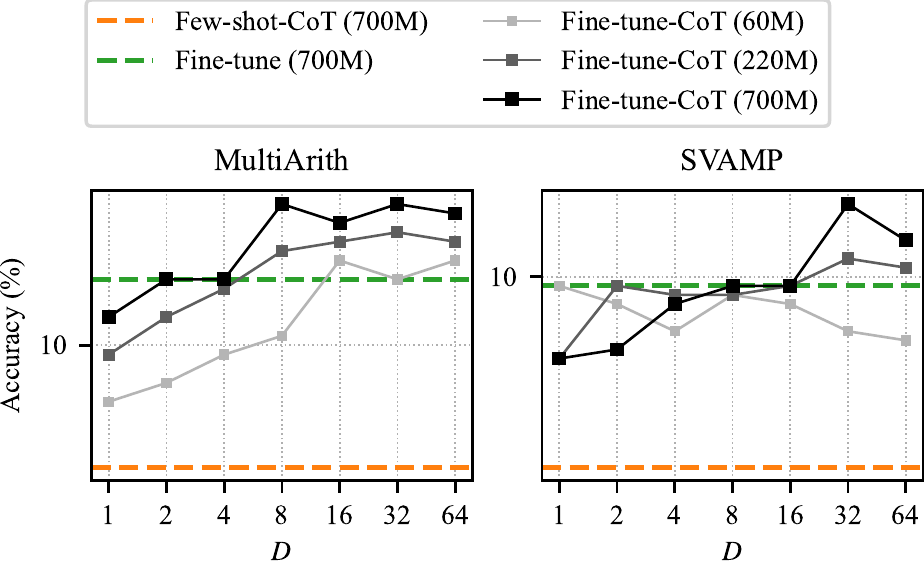}
        \caption{T5}
        \label{fig:diverse_reasoning_t5}
    \end{subfigure} \\
    \vspace{0.15cm}
    \begin{subfigure}{0.45\textwidth}
        \includegraphics[width=\textwidth]{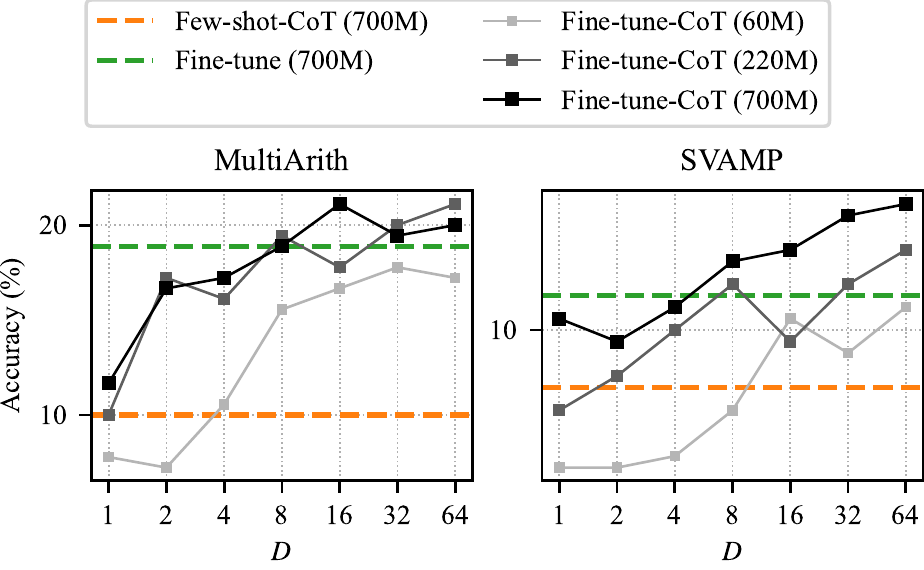}
        \caption{Flan-T5}
        \label{fig:diverse_reasoning_flan_t5}
    \end{subfigure} \\
    \vspace{0.15cm}
    \begin{subfigure}{0.45\textwidth}
        \includegraphics[width=\textwidth]{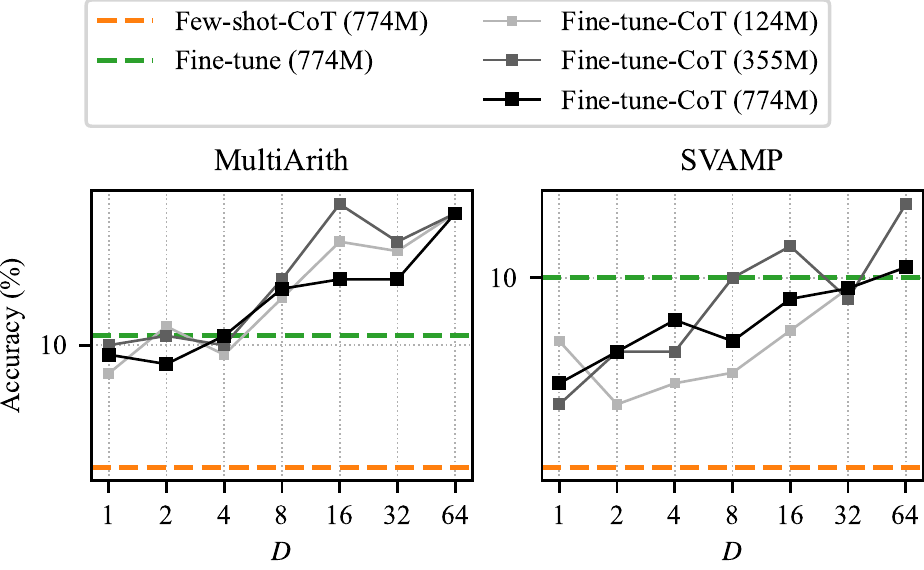}
        \caption{GPT-2}
        \label{fig:diverse_reasoning_gpt2}
    \end{subfigure}
    \caption{\textbf{Diverse reasoning performance on all models.} Accuracy (\%) of all student models under Fine-tune-CoT with varying degrees of diverse reasoning $D$. Baseline performance of the \textit{largest model} under vanilla fine-tuning and Few-shot-CoT are shown for comparison. Diverse reasoning is not applicable to the baselines.}
    \label{fig:diverse_reasoning_all}
\end{figure}

Figure~\ref{fig:diverse_reasoning_all} shows the performance of all student models on MultiArith and SVAMP under varying degrees of diverse reasoning.
We observe that performance scales with diverse reasoning in all student models, with the exception of T5-Small.
It is shown that diverse reasoning enables Fine-tune-CoT to outperform standard fine-tuning in all cases.

\paragraph{Effects of student model scale}
\label{student_model_size_open_source}

\begin{figure}[p]
    \centering
    \begin{subfigure}{0.45\textwidth}
        \includegraphics[width=\textwidth]{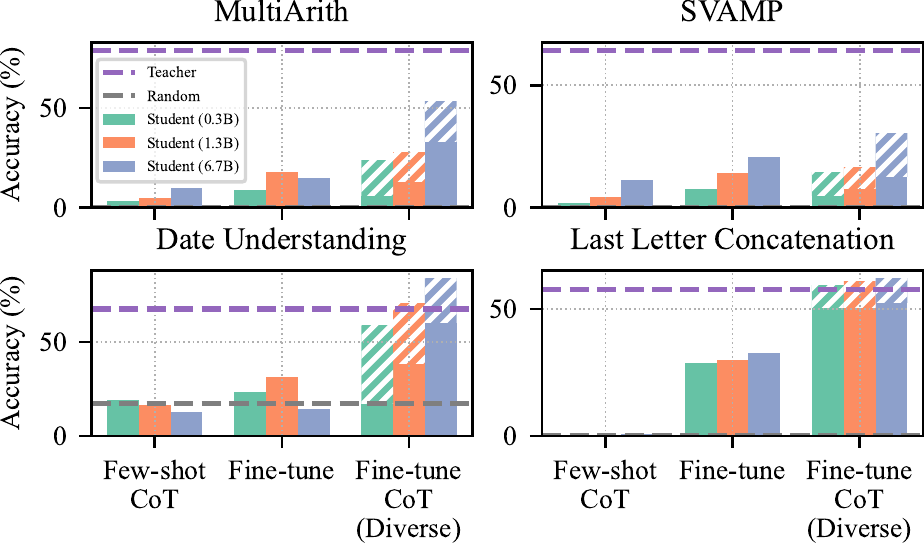}
        \caption{GPT-3}
        \label{fig:student_size_gpt3}
    \end{subfigure} \\
    \vspace{0.25cm}
    \begin{subfigure}{0.45\textwidth}
        \includegraphics[width=\textwidth]{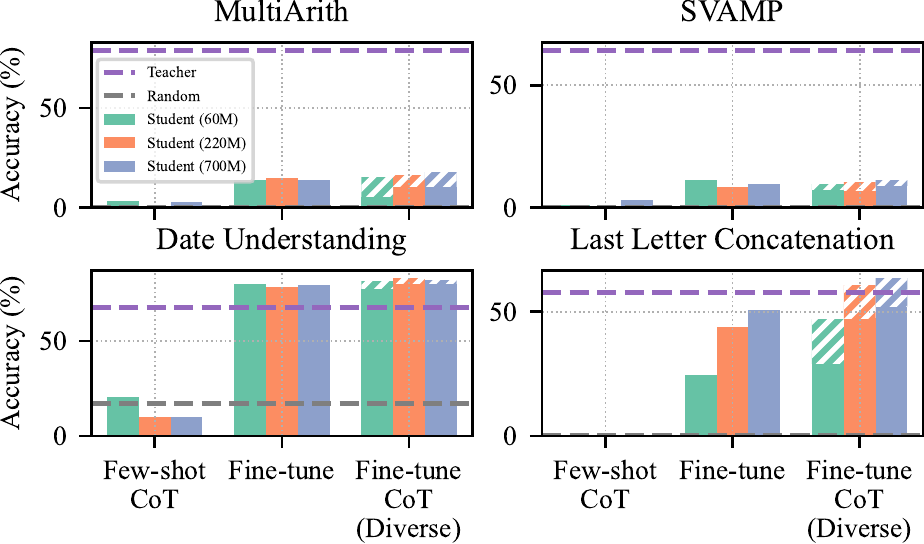}
        \caption{T5}
        \label{fig:student_size_t5}
    \end{subfigure} \\
    \vspace{0.25cm}
    \begin{subfigure}{0.45\textwidth}
        \includegraphics[width=\textwidth]{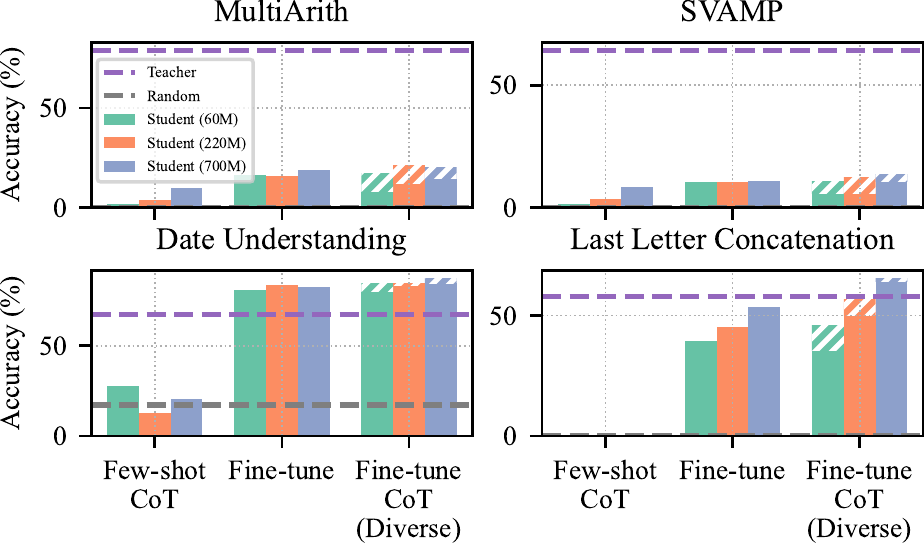}
        \caption{Flan-T5}
        \label{fig:student_size_flan_t5}
    \end{subfigure} \\
    \vspace{0.25cm}
    \begin{subfigure}{0.45\textwidth}
        \includegraphics[width=\textwidth]{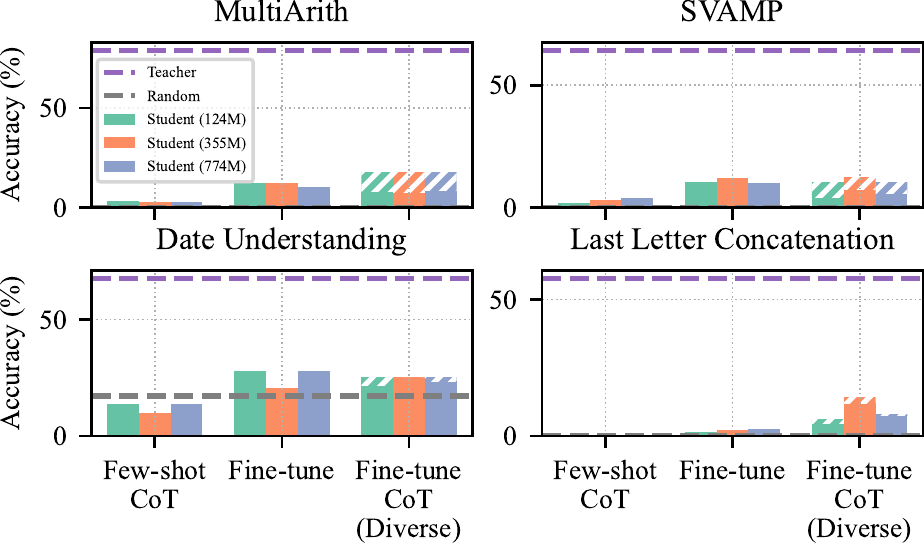}
        \caption{GPT-2}
        \label{fig:student_size_gpt2}
    \end{subfigure}
    \caption{\textbf{Effects of student model scale on all models.} Accuracy (\%) of all student models of various sizes under Few-shot-CoT, vanilla fine-tuning, and Fine-tune-CoT (with diverse reasoning). The hatched portion indicates the performance boost of Fine-tune-CoT when using diverse reasoning with $D=64$ on MultiArith and SVAMP; $D=8$ on others.}
    \label{fig:student_size_all}
\end{figure}

Figure~\ref{fig:student_size_all} shows the performance of all student model families according to model size.
While we observe performance scaling for Fine-tune-CoT on GPT-3 models, this is not apparent in other open-source models.
We posit that this may be due to under-tuned hyperparameters, as we used fixed hyperparameters for all open-source models, in contrast to default suggested settings for GPT-3.

\begin{table*}
\begin{minipage}[t]{1.0\linewidth}
\centering
\resizebox{\linewidth}{!}{
\begin{tabular}{llllllllllllll}
\toprule
\multirow[c]{2}{*}{Method} & \multirow[c]{2}{*}{Params} & Single\,\,\,\, & Add\,\,\,\,\,\, & Multi\,\,\,\, & \multirow[c]{2}{*}{GSM8K} & \multirow[c]{2}{*}{Aqua\,\,\,\,} & \multirow[c]{2}{*}{SVAMP} & Date & Shuffled & Last & Coin\,\,\,\, & Common & Strategy \\
& & Eq & Sub & Arith &  &  &  & Understanding & Objects & Letter & Flip & SenseQA & QA  \\
\midrule
Random &  & 0.00 & 0.00 & 0.00 & 0.00 & 20.00 & 0.00 & 17.12 & 33.33 & 0.00 & 50.00 & 20.00 & 50.00 \\
\midrule
\multicolumn{14}{c}{\textbf{Teacher: InstructGPT 175B (\texttt{text-davinci-002})}} \\
\midrule
Zero-shot-CoT & 175B & 81.50 & 76.71 & 78.79 & 42.17 & 29.74 & 64.20 & 67.58 & 53.20 & 57.71 & 90.04 & 60.07 & 53.45 \\
\midrule
\multicolumn{14}{c}{\textbf{Student: GPT-3 (\texttt{ada}, \texttt{babbage}, \texttt{curie})}}\\
\midrule
\multirow[t]{3}{*}{Few-shot-CoT} & 0.3B & 0.66 & 0.84 & 3.33 & 1.74 & 15.75 & 2.00 & 19.27 & - & 0.00 & 44.67 & 18.43 & 42.98 \\
 & 1.3B & 3.29 & 5.88 & 5.00 & 1.59 & 13.78 & 4.33 & 16.51 & - & 0.00 & 46.00 & 18.67 & 46.05 \\
 & 6.7B & 22.37 & 31.93 & 10.00 & 2.50 & 15.75 & 11.33 & 12.84 & - & 0.67 & 40.00 & 24.73 & 54.68 \\
\midrule
\multirow[t]{3}{*}{Fine-tune} & 0.3B & 9.87 & 8.40 & 8.89 & 5.08 & 24.41 & 7.67 & 23.42 & 32.44 & 28.67 & 100.00 & 51.68 & 60.41 \\
 & 1.3B & 11.84 & 17.65 & 17.78 & 5.38 & 21.26 & 14.33 & 31.53 & 30.22 & 30.00 & 100.00 & 70.93 & 60.70 \\
 & 6.7B & 24.34 & 25.21 & 15.00 & 6.14 & 15.35 & 20.67 & 14.41 & 33.78 & 32.67 & 72.00 & 76.17 & 65.21 \\
\midrule
\multirow[t]{3}{*}{Fine-tune-CoT} & 0.3B & 7.24 & 6.72 & 6.11 & 3.11 & 23.62 & 5.00 & 17.12 & 49.33 & 50.67 & 99.33 & 32.68 & 52.55 \\
 & 1.3B & 11.18 & 11.76 & 13.33 & 4.70 & 19.69 & 8.00 & 38.74 & 52.44 & 50.67 & 100.00 & 43.08 & 52.69 \\
 & 6.7B & 20.39 & 21.01 & 33.33 & 6.75 & 24.02 & 12.67 & 60.36 & 64.44 & 52.67 & 98.67 & 56.76 & 55.02 \\
\midrule
Fine-tune-CoT & 0.3B & 9.21 & 10.08 & 23.89 & - & - & 14.33 & 58.56 & 61.78 & 59.33 & 99.33 & - & 57.21 \\
\small{w/ diverse reasoning} & 1.3B & 18.42 & 19.33 & 27.78 & - & - & 16.33 & 70.27 & 72.00 & 60.67 & 100.00 & - & 57.06 \\
 & 6.7B & 24.34 & 31.09 & 53.33 & - & - & 30.33 & 83.78 & 73.33 & 62.00 & 100.00 & - & 58.22 \\
\midrule
\multicolumn{14}{c}{\textbf{Student: T5-\{Small, Base, Large\}}}\\
\midrule
Few-shot-CoT & 60M & 1.32 & 3.36 & 3.33 & 1.97 & 24.80 & 1.33 & 20.72 & - & 0.00 & 44.67 & 19.25 & 46.00 \\
 & 220M & 1.97 & 2.52 & 1.11 & 1.74 & 23.23 & 0.33 & 9.91 & - & 0.00 & 55.33 & 13.35 & 52.55 \\
 & 700M & 1.32 & 1.68 & 2.78 & 2.43 & 19.69 & 3.00 & 9.91 & - & 0.00 & 55.33 & 18.92 & 53.13 \\
\midrule
Fine-tune & 60M & 5.92 & 8.40 & 13.89 & 4.02 & 29.92 & 11.33 & 80.18 & 94.22 & 24.67 & 100.00 & 22.11 & 58.81 \\
 & 220M & 5.92 & 11.76 & 15.00 & 5.00 & 24.80 & 8.67 & 78.38 & 37.78 & 44.00 & 100.00 & 51.60 & 59.24 \\
 & 700M & 6.58 & 9.24 & 13.89 & 4.25 & 26.77 & 9.67 & 79.28 & 33.78 & 50.67 & 100.00 & 20.88 & 61.72 \\
\midrule
{Fine-tune-CoT} & 60M & 2.63 & 5.04 & 5.56 & 2.58 & 24.02 & 9.33 & 77.48 & 40.00 & 29.33 & 100.00 & 29.48 & 54.73 \\
& 220M & 4.61 & 7.56 & 10.56 & 3.18 & 26.77 & 7.00 & 80.18 & 42.67 & 47.33 & 98.67 & 45.37 & 55.90 \\
 & 700M & 5.26 & 10.92 & 10.56 & 4.55 & 29.92 & 9.00 & 80.18 & 46.22 & 52.00 & 100.00 & 54.22 & 56.33 \\
\midrule
{Fine-tune-CoT} & 60M & 7.24 & 7.56 & 15.00 & - & - & 7.67 & 81.08 & 59.11 & 46.67 & 100.00 & - & 56.04 \\
\small{w/ diverse reasoning} & 220M & 5.26 & 10.08 & 16.11 & - & - & 10.33 & 82.88 & 65.33 & 60.67 & 100.00 & - & 59.68 \\
 & 700M & 7.89 & 11.76 & 17.78 & - & - & 11.33 & 81.98 & 81.78 & 63.33 & 100.00 & - & 62.15 \\
\midrule
\multicolumn{14}{c}{\textbf{Student: Flan-T5-\{Small, Base, Large\}}} \\
\midrule
{Zero-shot} & 60M & 0.00 & 0.00 & 1.67 & 2.12 & 23.62 & 2.00 & 32.43 & 33.78 & 0.00 & 54.00 & 39.07 & 48.47 \\
 & 220M & 1.32 & 0.00 & 5.00 & 2.50 & 27.95 & 2.00 & 30.63 & 31.11 & 0.00 & 7.33 & 72.24 & 53.42 \\
 & 700M & 1.32 & 4.20 & 3.89 & 2.05 & 24.41 & 2.67 & 9.91 & 28.89 & 0.00 & 54.00 & 84.03 & 49.34 \\
\midrule
{Few-shot-CoT} & 60M & 1.32 & 0.84 & 1.67 & 2.81 & 20.87 & 1.67 & 27.93 & - & 0.00 & 44.67 & 11.79 & 51.97 \\
 & 220M & 2.63 & 0.84 & 3.89 & 3.64 & 24.80 & 3.67 & 12.61 & - & 0.00 & 44.67 & 70.27 & 53.86 \\
 & 700M & 12.50 & 10.08 & 10.00 & 7.51 & 23.23 & 8.33 & 20.72 & - & 0.00 & 44.67 & 83.87 & 65.21 \\
\midrule
{Fine-tune} & 60M & 7.24 & 9.24 & 16.67 & 4.93 & 28.74 & 10.33 & 81.08 & 33.78 & 39.33 & 100.00 & 45.95 & 58.95 \\
 & 220M & 5.26 & 10.08 & 16.11 & 5.08 & 29.53 & 10.67 & 83.78 & 44.00 & 45.33 & 100.00 & 63.55 & 61.14 \\
 & 700M & 7.24 & 12.61 & 18.89 & 5.53 & 24.80 & 11.00 & 82.88 & 33.78 & 53.33 & 100.00 & 66.75 & 63.90 \\
\midrule
{Fine-tune-CoT} & 60M & 6.58 & 5.88 & 8.33 & 2.96 & 23.23 & 5.67 & 80.18 & 36.00 & 35.33 & 100.00 & 42.01 & 54.15 \\
 & 220M & 4.61 & 9.24 & 12.22 & 4.40 & 29.13 & 6.00 & 83.78 & 48.89 & 50.00 & 100.00 & 59.05 & 59.97 \\
 & 700M & 11.84 & 10.92 & 14.44 & 5.38 & 28.35 & 10.67 & 84.68 & 55.11 & 64.00 & 100.00 & 66.83 & 59.83 \\
\midrule
{Fine-tune-CoT} & 60M & 7.24 & 10.92 & 17.22 & - & - & 10.67 & 84.68 & 62.22 & 46.00 & 100.00 & - & 56.04 \\
\small{w/ diverse reasoning} & 220M & 9.21 & 10.92 & 21.11 & - & - & 12.33 & 84.68 & 67.11 & 56.67 & 100.00 & - & 60.84 \\
 & 700M & 10.53 & 15.13 & 20.00 & - & - & 13.67 & 87.39 & 89.33 & 65.33 & 100.00 & - & 61.72 \\
\midrule
\multicolumn{14}{c}{\textbf{GPT-2 \{Small, Medium, Large\}}} \\
\midrule
{Few-shot-CoT} & 124M & 1.32 & 0.00 & 0.00 & 0.45 & 17.32 & 0.33 & 13.51 & - & 0.00 & 44.67 & 20.15 & 0.00 \\
 & 355M & 0.00 & 0.00 & 0.56 & 0.00 & 3.94 & 0.00 & 9.91 & - & 0.00 & 55.33 & 0.00 & 0.15 \\
 & 774M & 0.00 & 0.00 & 0.00 & 0.00 & 0.39 & 0.00 & 13.51 & - & 0.00 & 55.33 & 0.16 & 35.08 \\
\midrule
{Fine-tune} & 124M & 2.63 & 3.36 & 11.67 & 2.88 & 25.59 & 7.67 & 7.21 & 33.78 & 0.67 & 60.00 & 20.80 & 54.00 \\
 & 355M & 0.66 & 0.84 & 5.00 & 0.38 & 18.90 & 0.00 & 23.42 & 36.89 & 1.33 & 57.33 & 19.82 & 50.22 \\
 & 774M & 1.32 & 5.04 & 8.33 & 2.58 & 24.80 & 7.67 & 13.51 & 32.44 & 0.67 & 1.33 & 20.88 & 53.57 \\
\midrule
{Fine-tune-CoT} & 124M & 4.61 & 4.20 & 10.00 & 3.03 & 24.02 & 5.67 & 17.12 & 38.67 & 4.67 & 88.00 & 22.19 & 53.57 \\
 & 355M & 3.29 & 5.88 & 7.22 & 2.73 & 23.62 & 7.33 & 28.83 & 35.56 & 10.67 & 80.00 & 22.03 & 55.02 \\
 & 774M & 3.95 & 5.88 & 10.56 & 2.58 & 22.05 & 6.33 & 15.32 & 39.11 & 4.00 & 89.33 & 25.80 & 53.13 \\
\midrule
{Fine-tune-CoT} & 124M & 7.24 & 9.24 & 19.44 & - & - & 10.67 & 21.62 & 57.33 & 10.67 & 93.33 & - & 56.62 \\
\small{w/ diverse reasoning} & 355M & 5.92 & 9.24 & 17.22 & - & - & 9.67 & 20.72 & 56.00 & 20.00 & 95.33 & - & 55.60 \\
 & 774M & 8.55 & 12.61 & 17.22 & - & - & 8.67 & 18.02 & 52.44 & 7.33 & 84.67 & - & 57.06 \\
\bottomrule
\end{tabular}
}
\end{minipage}
\caption{
    \textbf{Fine-tune-CoT performance on all models.} Accuracy (\%) of all models on 12 tasks under Fine-tune-CoT (with diverse reasoning) and baseline methods.
   `Random' refers to random-guess performance derived based on the number of choices in multi-choice tasks.
   For diverse reasoning, we report results for maximum degree $D$ considered: $D=64$ for MultiArith and SVAMP; $D=8$ for other datasets.
   We omit diverse reasoning for large datasets due to resource constraints and Few-shot-CoT for Tracking Shuffled Objects due to absence of prompts.
   Zero-shot baseline performance is omitted due to negligible performance, except for Flan-T5 models.
}
\label{tab:performance_all}
\end{table*}

\clearpage

\clearpage

\begin{table*}
\begin{minipage}[b]{1.0\linewidth}
\caption{
    Examples of incorrect output.
}
\label{table:ex1}
\centering
\resizebox{\linewidth}{!}{
\begin{tabular}{| m{5em} | m{15cm}|} 
\toprule
\multicolumn{2}{c}{{\bf (a) GSM8K}}  \\
\midrule
\textbf{Question} &Zaid spends 1/4 of his salary on rent, 1/3 on car fuel and donates half of the remaining amount to his favorite charity. He gives his daughter 200\$ to use for her weekly expenses and 700\$ to his wife to budget for groceries and other household goods. If Zaid earns 6000\$ per month, how much money will he still have after all these expenses and donations? \texttt{\#\#\#} \\
\midrule
\textbf{Answer} & 350 \\
\midrule
\textbf{Completion} &Zaid spends 1/4 of his salary on rent, so that’s 300\$ per month. Zaid spends 1/3 of his salary on car fuel, so that’s 150\$ per month. That leaves him with 3000\$ per month. He donates half of the remaining amount to his favorite charity, so that’s 1500\$ per month. That leaves him with 1500\$ per month. He gives his daughter 200\$ per month, so that’s 400\$ per month. He gives his wife 700\$ per month, so that’s 350\$ per month.\\
\midrule
\textbf{Prediction}&350 \\
\bottomrule
\end{tabular}
}

\resizebox{\linewidth}{!}{
\begin{tabular}{| m{5em} | m{15cm}|} 
\toprule
\multicolumn{2}{c}{{\bf (b) AQUA}}  \\
\midrule
\textbf{Question} &20 marbles were pulled out of a bag of only white marbles, painted black, and then put back in. Then, another 20 marbles were pulled out, of which 1 was black, after which they were all returned to the bag. If the percentage of black marbles pulled out the second time represents their percentage in the bag, how many marbles in total Q does the bag currently hold? Answer choices: (A) 40, (B) 200, (C) 380, (D) 400, (E) 3200. \texttt{\#\#\#} \\
\midrule
\textbf{Answer} & D \\
\midrule
\textbf{Completion} & There are 20 white marbles, 20 black marbles, and 1 black marble. The first time, 10 white marbles and 10 black marbles are pulled out. The second time, 5 white marbles and 5 black marbles are pulled out. Now, we can calculate the total number of marbles that are in the bag. There are 20 white marbles and 20 black marbles. The first time, 10 white marbles and 10 black marbles are pulled out. The second time, 5 white marbles and 5 black \texttt{-}\texttt{-}\texttt{>} D \\
\midrule
\textbf{Prediction}&D \\
\bottomrule
\end{tabular}
}

\resizebox{\linewidth}{!}{
\begin{tabular}{| m{5em} | m{15cm}|} 
\toprule
\multicolumn{2}{c}{{\bf (c)  StrategyQA}}  \\
\midrule
\textbf{Question} & Is it common to see frost during some college commencements?  \texttt{\#\#\#} \\
\midrule
\textbf{Answer} & Yes \\
\midrule 
\textbf{Completion} & First, what is a college commencement? A college commencement is the ceremony at which a student graduates from college. Second, what is frost? Frost is a type of weather that occurs when the temperature is below freezing and the ground is frozen. So, is it common to see frost during some college commencements? No, it is not common to see frost during some college commencements. \texttt{-}\texttt{-}\texttt{>} No \\
\midrule
\textbf{Prediction}& No\\
\bottomrule
\end{tabular}
}
\end{minipage}
\end{table*}

\begin{table*}
\begin{minipage}[b]{1.0\linewidth}
\caption{
    Examples of incorrect output.
}
\label{table:ex2}
\centering

\resizebox{\linewidth}{!}{
\begin{tabular}{| m{5em} | m{15cm}|} 
\toprule
\multicolumn{2}{c}{{\bf (a) StrategyQA}}  \\
\midrule
\textbf{Question} &Is sternum connected to any of the bones mentioned in James Weldon Johnson’s Dem Bones?  \texttt{\#\#\#} \\
\midrule
\textbf{Answer} & Yes \\
\midrule
\textbf{Completion} &The sternum is a bone in the chest. The bones in the chest are the sternum, the clavicle, the two collarbones, and the two breastbones. The breastbones are the two right and left breastbones. So, no, the sternum is not connected to any of the bones mentioned in James Weldon Johnson’s Dem Bones. \texttt{-}\texttt{-}\texttt{>} No\\
\midrule
\textbf{Prediction}&No \\
\bottomrule
\end{tabular}
}

\resizebox{\linewidth}{!}{
\begin{tabular}{| m{5em} | m{15cm}|} 
\toprule
\multicolumn{2}{c}{{\bf (b) SingleEq}}  \\
\midrule
\textbf{Question} &There are 33 walnut trees currently in the park. Park workers will plant 44 more walnut trees today. How many walnut trees will the park have when the workers are finished ? \texttt{\#\#\#} \\
\midrule
\textbf{Answer} & 77.0 \\
\midrule
\textbf{Completion} & There are 33 walnut trees currently in the park. Park workers will plant 44 more walnut trees today. This means that when the workers are finished, there will be a total of 83 walnut trees in the park. \texttt{-}\texttt{-}\texttt{>} 83.0\\
\midrule
\textbf{Prediction}&83.0 \\
\bottomrule
\end{tabular}
}

\resizebox{\linewidth}{!}{
\begin{tabular}{| m{5em} | m{15cm}|} 
\toprule
\multicolumn{2}{c}{{\bf (c) AddSub 25}}  \\
\midrule
\textbf{Question} & Jason has 43 blue and 16 red marbles . Tom has 24 blue marbles . How many blue marbles do they have in all ?\texttt{\#\#\#} \\
\midrule
\textbf{Answer} & 67.0\\
\midrule
\textbf{Completion} &  Jason has 43 blue marbles and 16 red marbles. Tom has 24 blue marbles. To find out how many blue marbles they have in all, we need to add 43 and 24. 43 + 24 = 77 Therefore, they have 77 blue marbles in all. \texttt{-}\texttt{-}\texttt{>} 77\\
\midrule
\textbf{Prediction}&77 \\
\bottomrule
\end{tabular}
}

\resizebox{\linewidth}{!}{
\begin{tabular}{| m{5em} | m{15cm}|} 
\toprule
\multicolumn{2}{c}{{\bf (d) GSM8K}}  \\
\midrule
\textbf{Question} &  Four dogs sat in a line within the veterinarian’s waiting room. The first dog was the Affenpinscher, who weighed only 10 pounds. Next to him sat a Dachshund, who weighed twice as much as the Affenpinscher. Next to the Dachshund sat a Papillon, who weighed one-fourth as much as the Dachshund. And at the end of the line sat a Mastiff, who weighed 44 times the weight of the Papillon. How much did the Mastiff weigh, in pounds? Let’s think step by step \texttt{\#\#\#} \\
\midrule
\textbf{Answer} & 220 \\
\midrule 
\textbf{Completion} &The Affenpinscher weighs 10 pounds. The Dachshund weighs twice as much as the Affenpinscher, so it weighs 20 pounds. The Papillon weighs one-fourth as much as the Dachshund, so it weighs 2 pounds. The Mastiff weighs 44 times as much as the Papillon, so it weighs 44 pounds. \texttt{-}\texttt{-}\texttt{>} 44\\
\midrule
\textbf{Prediction}&44 \\
\bottomrule
\end{tabular}
}
\end{minipage}
\end{table*}

\begin{table*}
\begin{minipage}[b]{1.0\linewidth}
\caption{
    Examples of incorrect output.
}
\label{table:ex3}
\centering
\resizebox{\linewidth}{!}{
\begin{tabular}{| m{5em} | m{15cm}|} 
\toprule
\multicolumn{2}{c}{{\bf (a) SVAMP}}  \\
\midrule
\textbf{Question} & If you had 33 cookies and each bag has 11 cookies How many bags of cookies do you have? \texttt{\#\#\#} \\
\midrule
\textbf{Answer} & 3.0 \\
\midrule
\textbf{Completion} & There are 33 cookies. Each bag has 11 cookies. That means that if we divide 33 by 11, we will know how many bags of cookies there are. 33 divided by 11 equals 2. There are 2 bags of cookies. \texttt{-}\texttt{-}\texttt{>} 2.0 \\
\midrule
\textbf{Prediction}&2.0 \\
\bottomrule
\end{tabular}
}

\resizebox{\linewidth}{!}{
\begin{tabular}{| m{5em} | m{15cm}|} 
\toprule
\multicolumn{2}{c}{{\bf (b) CommonsenseQA} }  \\
\midrule
\textbf{Question} & What is someone who isn’t clever, bright, or competent called? Answer choices: (A) clumsy, (B) ineffectual, (C) dull, (D) clumsy, (E) stupid. \texttt{\#\#\#} \\
\midrule
\textbf{Answer} & E \\
\midrule
\textbf{Completion} & If someone isn’t clever, bright, or competent, then they’re not effective. If someone’s not effective, then they’re not useful. If someone’s not useful, then they’re not needed. If someone’s not needed, then they’re not important. Therefore, the answer is (B) ineffectual. \texttt{-}\texttt{-}\texttt{>} B\\
\midrule
\textbf{Prediction}& B \\
\bottomrule
\end{tabular}
}

\resizebox{\linewidth}{!}{
\begin{tabular}{| m{5em} | m{15cm}|} 
\toprule
\multicolumn{2}{c}{{\bf (c) CommonsenseQA }}  \\
\midrule
\textbf{Question} &  Billy set aside a block of time for having fun after work. Why might he do this? Answer choices: (A) happiness, (B) stress relief, (C) pleasure, (D) ocean, (E) may laugh. \texttt{\#\#\#} \\
\midrule
\textbf{Answer} & B \\
\midrule
\textbf{Completion} & After work, Billy might be tired and want to relax. He could set aside a block of time and say, “OK, I’m going to spend the next X minutes doing something I enjoy.” This might help him relax and feel happier. \texttt{-}\texttt{-}\texttt{>} A\\
\midrule
\textbf{Prediction}&A \\
\bottomrule
\end{tabular}
}

\resizebox{\linewidth}{!}{
\begin{tabular}{| m{5em} | m{15cm}|} 
\toprule
\multicolumn{2}{c}{{\bf (d) CommonsenseQA}}  \\
\midrule
\textbf{Question} &  What island country is ferret popular? Answer choices: (A) own home, (B) north carolina, (C) great britain, (D) hutch, (E) outdoors.\texttt{\#\#\#} \\
\midrule
\textbf{Answer} & C \\
\midrule
\textbf{Completion} & A ferret is a small mammal, so (D) hutch is probably not the best answer. It's also not a very common animal, so (A) own home and (B) north carolina are probably not the best answers. That leaves us with (C) great britain and (E) outdoors. Great britain is a larger island than outdoors, so the ferret is probably more popular in that country. \texttt{-}\texttt{-}\texttt{>} C	 \\
\midrule
\textbf{Prediction}&C \\
\bottomrule
\end{tabular}
}

\resizebox{\linewidth}{!}{
\begin{tabular}{| m{5em} | m{15cm}|} 
\toprule
\multicolumn{2}{c}{{\bf (e) SVAMP}}  \\
\midrule
\textbf{Question} & There are 17 different movies and 11 different books in the ’ crazy silly school ’ series. If you read 13 of the books and watched 63 of the movies How many more movies than books are there in the ’ crazy silly school ’ series? \texttt{\#\#\#} \\
\midrule
\textbf{Answer} & 6.0 \\
\midrule
\textbf{Completion} &There are 17 different movies and 11 different books in the ‘crazy silly school’ series. If you read 13 of the books, that means you have read 7 books. If you watched 63 of the movies, that means you have watched 63 movies. So there are 17 movies and 11 books, and you have read 7 books and watched 63 movies. That means there are 8 more movies than books in the ‘crazy silly school’ series. \texttt{-}\texttt{-}\texttt{>} 8.0\\
\midrule
\textbf{Prediction}&8.0 \\
\bottomrule
\end{tabular}
}
\end{minipage}
\end{table*}

\begin{table*}
\begin{minipage}[b]{1.0\linewidth}
\caption{
    Examples of incorrect output.
}
\label{table:ex4}
\centering
\resizebox{\linewidth}{!}{
\begin{tabular}{| m{5em} | m{15cm}|} 
\toprule
\multicolumn{2}{c}{{\bf (a) SVAMP }}  \\ 
\midrule
\textbf{Question} & Danny collects bottle caps and wrappers. He found 65 wrappers and 5 bottle caps at the park. Now he has 31 bottle caps and 9 wrappers in his collection. How many more wrappers than bottle caps did danny find at the park? \texttt{\#\#\#} \\
\midrule
\textbf{Answer} & 60.0 \\
\midrule
\textbf{Completion} & First, Danny finds 65 wrappers and 5 bottle caps at the park. Then, he has 31 bottle caps and 9 wrappers in his collection. Now, we want to know how many more wrappers than bottle caps Danny found at the park. We can find this out by subtracting the number of bottle caps from the number of wrappers: 9 - 5 = 4 So, Danny found 4 more wrappers than bottle caps at the park. \texttt{-}\texttt{-}\texttt{>} 4.0\\
\midrule
\textbf{Prediction}&4.0 \\
\bottomrule
\end{tabular}
}

\resizebox{\linewidth}{!}{
\begin{tabular}{| m{5em} | m{15cm}|} 
\toprule
\multicolumn{2}{c}{{\bf (b) SVAMP }}  \\ 
\midrule
\textbf{Question} &  Next on his checklist is wax to stick the feathers together and colors to paint them. He needs a total of 57 g of colors to paint them. He needs 22 g of wax more. If the feathers require a total of 353 g of wax How many grams of wax does he already have? \texttt{\#\#\#} \\
\midrule
\textbf{Answer} & 331.0 \\
\midrule
\textbf{Completion} & First, he needs 57 grams of colors to paint the feathers. Second, he needs 22 grams of wax more. Third, the feathers require a total of 353 grams of wax. So, he already has 353 grams of wax. \texttt{-}\texttt{-}\texttt{>} 353.0\\
\midrule
\textbf{Prediction}&353.0 \\
\bottomrule
\end{tabular}
}

\resizebox{\linewidth}{!}{
\begin{tabular}{| m{5em} | m{15cm}|} 
\toprule
\multicolumn{2}{c}{{\bf (c) CommonsenseQA }}  \\ 
\midrule
\textbf{Question} & You’ll find a landing at the top of what? Answer choices: (A) ocean, (B) apartment building, (C) stairwell, (D) airport, (E) room. \texttt{\#\#\#} \\
\midrule
\textbf{Answer} & C \\
\midrule
\textbf{Completion} &A landing is a place where something can touch the ground. A landing at the top of an ocean would be a very strange place to find, since there would be no ground to touch. A landing at the top of an apartment building would be a very strange place to find, since there would be no ground to touch. A landing at the top of a stairwell would be a very strange place to find, since there would be no ground to touch. A landing at the top of an airport would be a very strange place to find, since there would be no ground to\\
\midrule
\textbf{Prediction}&A \\
\bottomrule
\end{tabular}
}

\resizebox{\linewidth}{!}{
\begin{tabular}{| m{5em} | m{15cm}|} 
\toprule
\multicolumn{2}{c}{{\bf (d) SVAMP }}  \\ 
\midrule
\textbf{Question} &  18 red peaches, 14 yellow peaches and 17 green peaches are in the basket. How many red and yellow peaches are in the basket?\texttt{\#\#\#} \\
\midrule
\textbf{Answer} & 32.0 \\
\midrule
\textbf{Completion} & There are a total of 48 peaches in the basket. There are 18 red peaches, 14 yellow peaches, and 17 green peaches. This means that there are a total of 33 peaches in the basket. There are 33 peaches in the basket. This means that there are 17 red peaches and 14 yellow peaches in the basket. \texttt{-}\texttt{-}\texttt{>} 17.0 \\
\midrule
\textbf{Prediction}&17.0 \\
\bottomrule
\end{tabular}
}
\end{minipage}
\end{table*}

\begin{table*}
\begin{minipage}[b]{1.0\linewidth}
\caption{
    Examples of correct output.
}
\label{table:ex5}
\centering
\resizebox{\linewidth}{!}{
\begin{tabular}{| m{5em} | m{15cm}|} 
\toprule
\multicolumn{2}{c}{{\bf (a) Date Understanding }}  \\  
\midrule
\textbf{Question} &Jane was born on the last day of Feburary in 2000. Today is her 16-year-old birthday. What is the date a month ago in MM/DD/YYYY? Which choice is true? Answer choices: (A) 01/30/2016, (B) 11/08/2015, (C) 01/29/2016, (D) 07/29/2015, (E) 05/04/2016, (F) 02/02/2016. \texttt{\#\#\#} \\
\midrule
\textbf{Answer} & C \\
\midrule
\textbf{Completion} & Jane’s 16th birthday is on 02/28/2016. A month ago from that would be 01/28/2016. Therefore, the answer is (C). \texttt{-}\texttt{-}\texttt{>} C\\
\midrule
\textbf{Prediction}&C \\
\bottomrule
\end{tabular}
}

\resizebox{\linewidth}{!}{
\begin{tabular}{| m{5em} | m{15cm}|} 
\toprule
\multicolumn{2}{c}{{\bf (b) Last Letter Concatenation }}  \\  
\midrule
\textbf{Question} & Take the last letters of each words in "Lino Mariel Aditya Elisabeth" and concatenate them. \texttt{\#\#\#} \\
\midrule
\textbf{Answer} & olah \\
\midrule
\textbf{Completion} &The last letter of "Lino" is "o". The last letter of "Mariel" is "l". The last letter of "Aditya" is "a". The last letter of "Elisabeth" is "h". So the final answer is "olah". --> olah\\
\midrule
\textbf{Prediction}&olah \\
\bottomrule
\end{tabular}
}

\resizebox{\linewidth}{!}{
\begin{tabular}{| m{5em} | m{15cm}|} 
\toprule
\multicolumn{2}{c}{{\bf (c) Coin Flip }}  \\ 
\midrule
\textbf{Question} &A coin is heads up. Cinthia does not flip the coin. Lloyd does not flip the coin. Jacqueline flips the coin. Jc flips the coin. Is the coin still heads up? Note that "flip" here means "reverse". \texttt{\#\#\#} \\
\midrule
\textbf{Answer} & Yes \\
\midrule
\textbf{Completion} & Cinthia does not flip the coin, so the coin remains heads up. Lloyd does not flip the coin, so the coin remains heads up. Jacqueline flips the coin, so the coin is now tails up. Jc flips the coin, so the coin is now heads up again. --> yes\\
\midrule
\textbf{Prediction}& Yes \\
\bottomrule
\end{tabular}
}

\resizebox{\linewidth}{!}{
\begin{tabular}{| m{5em} | m{15cm}|} 
\toprule
\multicolumn{2}{c}{{\bf (d) Tracking Shuffled Objects }}  \\ 
\midrule
\textbf{Question} &  Alice, Bob, and Claire are playing a game. At the start of the game, they are each holding a ball: Alice has a blue ball, Bob has a yellow ball, and Claire has a green ball. As the game progresses, pairs of players trade balls. First, Bob and Claire swap balls. Then, Bob and Alice swap balls. Finally, Bob and Claire swap balls. At the end of the game, Bob has the Which choice is true? Answer choices: (A) blue ball, (B) yellow ball, (C) green ball. \texttt{\#\#\#} \\
\midrule
\textbf{Answer} & B \\
\midrule
\textbf{Completion} & First, Bob and Claire swap balls. This means that Bob now has the green ball and Claire has the yellow ball. Then, Bob and Alice swap balls. This means that Bob now has the blue ball and Alice has the green ball. Finally, Bob and Claire swap balls again. This means that Bob now has the yellow ball and Claire has the green ball. At the end of the game, Bob has the yellow ball. --> B	 \\
\midrule
\textbf{Prediction}&B \\
\bottomrule
\end{tabular}
}

\resizebox{\linewidth}{!}{
\begin{tabular}{| m{5em} | m{15cm}|} 
\toprule
\multicolumn{2}{c}{{\bf (e) Date Understanding } }  \\ 
\midrule
\textbf{Question} & Tomorrow is 11/12/2019. What is the date one year ago from today in MM/DD/YYYY? Which choice is true? Answer choices: (A) 11/02/2018, (B) 11/04/2018, (C) 09/04/2018, (D) 08/25/2018, (E) 11/11/2018. \texttt{\#\#\#} \\
\midrule
\textbf{Answer} & E \\
\midrule
\textbf{Completion} & One year ago from today would be 11/11/2018. \texttt{-}\texttt{-}\texttt{>} E\\
\midrule
\textbf{Prediction}& E \\
\bottomrule
\end{tabular}
}
\end{minipage}
\end{table*}